\documentclass[ds,nolinenumbering]{mscthesis}

\begin{document}
\pagestyle{plain}
\setcounter{page}{0}
\title{Predicting Delayed Trajectories Using Network Features: A Study on the Dutch Railway Network}

\author{Merel Kämpere$^{1}$, Ali Mohammed Mansoor Alsahag$^{1}$}
\affiliation{%
  \institution{$^{1}$Informatics Institute, University of Amsterdam}
  \streetaddress{1098XH Science Park}
  \city{Amsterdam}
  \country{The Netherlands}
}
\email{merel.kamper@student.uva.nl, a.m.m.alsahag@uva.nl}

\begin{abstract}

The Dutch railway network is one of the busiest in the world, with delays being a prominent concern for the principal passenger railway operator NS. This research addresses a gap in delay prediction studies within the Dutch railway network by employing an XGBoost Classifier with a focus on topological features. Current research predominantly emphasizes short-term predictions and neglects the broader network-wide patterns essential for mitigating ripple effects. This research implements and improves an existing methodology, originally designed to forecast the evolution of the fast-changing US air network, to predict delays in the Dutch Railways. By integrating Node Centrality Measures and comparing multiple classifiers like RandomForest, DecisionTree, GradientBoosting, AdaBoost, and LogisticRegression, the goal is to predict delayed trajectories. However, the results reveal limited performance, especially in non-simultaneous testing scenarios, suggesting the necessity for more context-specific adaptations. Regardless, this research contributes to the understanding of transportation network evaluation and proposes future directions for developing more robust predictive models for delays.
\end{abstract}

\maketitle
\section{Introduction}
\label{sec:introduction}

Recently, the application of machine learning techniques to forecast delays \cite{li-2020, xu-2022} and optimize transportation networks \cite{yin-2023} has gained significant attention in academia and industry. These efforts are driven by the need to enhance transportation systems' efficiency, reliability, and sustainability \cite{colin-2016, ji-2022, stamos-2023}. With the power of data-driven approaches \cite{spanninger-2022}, researchers and stakeholders seek to understand the complex dynamics of transportation networks better and develop effective strategies for mitigating delays and improving overall network performance. Improving punctuality is also high on the agenda at the principal passenger railway operator in the Netherlands (NS) after it was announced that 1 in 10 journeys were delayed in 2023 \cite{dec-2023}.

Despite significant advancements in machine learning for predicting delays in transportation networks, recent studies have mostly focused on short-term predictions at the next station, typically within a 20-minute timeframe \cite{li-2020, xu-2022} or optimization of train dispatching \cite{zhang-2022, mou-2019, shi-2021}. These approaches overlook the broader network-wide delay patterns, which are crucial for understanding and mitigating ripple effects throughout the system \cite{huang-2024}. This research aims to address this gap by implementing and improving an existing machine-learning framework that focuses on the predictive power of topological features in forecasting the evolution of transportation networks to predict delays in the Dutch railway network. By emphasizing the significance of these network-related features in the prediction of delayed links, this research seeks to develop more comprehensive models that can capture complex interactions on the network level instead of short-term travels.

In transportation network forecasting, Lei et al. \cite{lei-2022} have made a significant impact regarding the evolution of the fast-changing networks in the US air transport system and the Brazilian bus network. Their study primarily focuses on predicting removed links over a monthly time scale, a methodology that fits the dynamic nature of fast-changing networks. This research seeks to expand the applicability of their framework to the Dutch railway network operated by NS. Unlike the fast-paced nature of air and bus transport, the Dutch railway network exhibits a slower rate of change, with delays being a prominent concern. Therefore, this research challenges the notion that Lei et al.'s methodology is exclusively suitable for fast-changing networks and predicting removed links. By redefining 'removed links' in their study to 'delayed links' in this research, the goal is to implement their framework to better align with the operational dynamics of the Dutch railway network. This implementation and improvement will enable the investigation of the generalizability of their approach in the Dutch railway network and contribute to a broader understanding of transportation network forecasting methodologies. The redefinition of the missing link prediction is justified by the fact that, when examining delayed links, this can still be encountered as a binary classification problem. Here, the objective shifts from determining retained or removed links to discerning whether an edge is (significantly) delayed or not within a specific timeframe. Moreover, a delayed link can be seen as a missing link, as when a delay occurs, passengers must seek alternative routes to continue their travels.

Contrary to prior studies that primarily examine the impact of spatiotemporal features on train delays \cite{huang-2020b, zhang-2022, wang-2024} or focus on predicting delays 20 minutes later \cite{li-2020, xu-2022}, this research emphasizes the identification of predictive network features that significantly influence delayed trajectories within the Dutch railway network. Inspired by Lei et al. \cite{lei-2022}, who highlighted the significance of topological features in predicting the evolution of transportation networks, this research shifts the focus from external variables to intrinsic network characteristics. By addressing this gap, the aim is to shed light on a less-explored aspect of delay prediction, offering valuable insights into vulnerable trajectories. When the location and cause of a delay are identified, potential new routes and stops can be analyzed. These insights could help in proactive decision-making processes and enable stakeholders to mitigate potential delays preemptively. According to B. van Zaalen (personal communication, June 4, 2024), head of Digitalisation Operations at NS, passenger punctuality is the most important operational KPI, next to seat probability. Punctuality for passengers on the main railway network was lower in 2023 than in 2022 with an average of 89.7\% \cite{unknown-author-no-date}. One of the reasons for the decline was an increase in train crowding after COVID. A lot of work is being done at NS to optimize operations and keep passenger punctuality as high as possible. Unfortunately, NS suffers a lot from a ripple effect: if one train is delayed or disrupted on a route, there is a big chance that other trains in that region will be as well. This research contributes to ongoing practices by investigating network-related features \cite{stamos-2023} as new predictors and emphasizing a scientific baseline.

Moreover, this research evaluates the effectiveness of various machine learning classifiers and feature sets by comparing them and enhancing a baseline model \cite{lei-2022}. The ultimate goal is to test the generalizability of the baseline work and improve it to predict delayed trajectories within the Dutch railway network. By employing these methodological combinations, this research not only contributes to the broader field of transportation network optimization but also strives to enhance the efficiency and reliability of the Dutch railway network. 
This study investigates the application and enhancement of machine learning techniques,particularly those traditionally used to forecast link removals in fast-changing networks,for predicting delayed train trajectories within the Dutch railway network operated by NS. The focus lies on evaluating the extent to which established models such as the XGBoost Classifier can be adapted and improved for this context. Furthermore, the research explores strategies for constructing more effective predictive models tailored to the specific characteristics of the Dutch railway system. The performance of these enhanced models is then benchmarked against baseline approaches to assess their effectiveness in forecasting delays, with particular attention given to the influence of key network features.

\section{Related Work}
\label{sec:related_work}

The prediction of public transport delays is a much-researched topic in the data science domain. Many different countries have been subject to this investigation, for example, Belgium \cite{sobrie-2023}, China \cite{shi-2021, huang-2020a}, India \cite{mohd-2021}, France \cite{lbazri-2020}, and Germany \cite{hauck-2020}. Besides that, different techniques have been used for network forecasts, of which many have been reviewed by making a distinction between event-driven and data-driven approaches \cite{spanninger-2022}. Furthermore, there is a focus on climate change when it concerns studies on transportation networks \cite{ji-2022, coles-2001}. In recent work, the usefulness and usability of centrality measures in transportation networks in the face of climate change adaptation have been evaluated \cite{stamos-2023}. Centrality measures are a powerful tool in network theory that can be used to understand the importance of nodes in a network. The review highlights the need to reformulate these measures, because when this is done, they can properly be applied in transportation networks to expose the significance of their elements. 

Numerous studies have also focused on predicting delays in the Dutch railway network, particularly concerning the primary provider NS. A significant incentive for this study stemmed from the \href{https://connect.informs.org/railway-applications/new-item3/problem-solving-competition681/new-item12}{Railroad Problem Solving Competition} of 2018. During this event, a sizable dataset from NS and ProRail, comprising timetables, weather data, and delay records, was made available. The objective was to enhance the accuracy of train performance and delay forecasts about 20 minutes after a realization. Notably, the top three finalists employed distinct models for delay estimation. The winning approach used a neural network \cite{haahr-2019}, the runner-up made use of a bi-level random forest method \cite{nabian-2019}, and the third finalist employed non-homogeneous Markov chains \cite{xu-2022}. At the primary level, the bi-level random forest predicts whether the current delay will decrease, increase, or remain unchanged within the next 20 minutes from the present time. At the secondary level, their model quantifies the amount of delay in minutes \cite{nabian-2019}. It is important to note that their model was significantly better at predicting the decrease of delay than the increase or equal delay. Other classifiers that were used in this study and had slightly lower accuracies than Random Forest were Gradient Boosting, SVM, Adaboost, Logistic Regression and Decision Tree. This Railroad Competition dataset has later been used by other researchers to predict near-term train delays \cite{xu-2022} or discover influencing factors for delay propagation \cite{li-2020}. While these studies show better results than their baseline models, their results are not optimal, and the limitations include the failure to consider other features. The dataset of these studies is used as inspiration for this research because it contains historical data on departures, arrivals, and delays.

The methodologies of this research build upon the work of Lei et al \cite{lei-2022}. Their study focuses on applying machine learning techniques to predict the evolution of dynamic transportation networks, specifically focusing on the US domestic air transport network and the domestic bus transport network in Brazil. A missing link prediction approach was used over a monthly time frame. The main steps in their work include comparing different classification models, testing if topological features are significant for removed edges compared to retained edges, studying the predictive potential of the topological features, testing resilience to external shocks such as the COVID-19 pandemic, and examining whether the forecast is stable over a longer period of time.

Lei et al. initially built graphs representing their transportation networks, with nodes denoting entities (airports or bus stations) and edges denoting connections (such as flights or bus routes). They extracted both weighted and unweighted topological features to capture the structural properties of their network. These features were assessed in terms of differences between retained edges and removed ones. The study tested 27 commonly used classification algorithms, including XGBoost, which emerged as the best performer based on balanced accuracy, F1 score, and ROC AUC. The resilience testing against external shocks, such as the impact of the COVID-19 pandemic, was incorporated by Lei et al. to understand the robustness of their models. They applied simultaneous and non-simultaneous testing to validate the models' ability to predict removed links both within the same time period and across different time periods. Furthermore, they used SHAP (SHapley Additive exPlanations) values to interpret the importance of features in their models, providing insights into which topological features were most influential in predicting link removals. Their results show that edge removal processes in transportation networks are not random and it is possible to make accurate predictions based on network structures. While simultaneous testing works well for both networks and the model can make accurate predictions in the same snapshot, a model trained on a single time snapshot is not able to correctly predict removed edges in different time snapshots for the Brazil Bus network. 

In conclusion, this research integrates the extensively studied topic of train delay predictions with the methodologies proposed by Lei et al. Their focus on the network as a whole and the analysis of topological features addresses a critical research gap in the context of the Dutch railway network. Implementing Lei et al.'s work, which utilized XGBoost for predicting the removal of links in a fast-changing transportation network, serves as a baseline and an important validation step \cite{naidu-2023} for the application to the NS dataset. Additionally, this research leverages insights from Spanninger et al. \cite{spanninger-2022} on various delay prediction methods, including event-driven and data-driven approaches. Specific classification models tailored to the NS data are evaluated for their effectiveness \cite{nabian-2019, vandebijl2022dutch}. Centrality measures are incorporated as indicators of network-wide importance, aiding in understanding delays \cite{stamos-2023}. Furthermore, this research adopts a similar approach for determining the arrival delay as a dependent variable compared to previous studies \cite{mou-2019, shi-2021, zhang-2022}, emphasizing consistency towards and expansion of the already existing literature in this field.

\section{Methodology}
\label{sec:methodology}
This section outlines the comprehensive methodology employed to predict delayed trajectories in the Dutch railway network using machine learning techniques from a baseline work where missing links are predicted in the US air network. The process begins with a description of the data collection and preparation stages of the NS dataset, followed by an exploratory data analysis to uncover patterns in both datasets. Subsequently, the model framework and improvements made to implement the work by Lei et al. are detailed. Finally, the validation and evaluation procedures are explained. Redoing the baseline work is an important part of the validation step of this research.

\subsection{Datasets} 
Two datasets are used in this research. Firstly, the US domestic flights as used by Lei et al. \cite{lei-2022}. This dataset is chosen over the Brazil Bus dataset because it showed better and more stable results in both testing scenarios. For this dataset, few cleaning and transformation steps needed to take place. Secondly, a dataset with train rides in the Netherlands is used. This dataset underwent more cleaning, preparation, and transformation steps to fit the baseline model and research approach. All of these efforts are described below, as well as the exploration of both datasets.

\subsubsection{Data collection} The US air data is publicly available and can be found referenced in the baseline work \cite{lei-2022} and in \href{https://arch.library.northwestern.edu/concern/datasets/pn89d700k}{the online library} created by authors of the baseline study. The original dataset contains the flight connections from January 2004 till March 2021, within the US (origin and destination details), the number of departures scheduled and performed, the number of seats, passengers, distance, carrier information on a monthly timescale. The original dataset can be reviewed in Table \ref{tab:flight_data1} in the Appendix.

The data utilized in this research is sourced from \href{https://www.rijdendetreinen.nl/en/open-data}{rijdendetreinen.nl} and is also publicly accessible. This data is derived from the real-time data provided by NS, including live departure times, live arrival times, and service updates. It is also used in the NS app and Rijden de Treinen's website, highlighting its reliability and relevance. The dataset contains all passenger train services in the Netherlands since 2019. While this exact dataset has not been widely used in research, similar datasets have demonstrated reliability in various studies \cite{besinovic-2020, szymula-2020, zhu-2018}. For instance, the disruptions dataset from the same website as the data used in this research has been employed in studies evaluating the resilience and vulnerability of transportation systems \cite{besinovic-2020, szymula-2020}, as well as in timetable rescheduling studies \cite{zhu-2018}. Furthermore, datasets used in the \href{https://connect.informs.org/railway-applications/new-item3/problem-solving-competition681/new-item12}{Railroad Problem Solving Competition} from 2018, provided by ProRail—the organization responsible for maintaining and extending the national railway network—share many similar attributes with the NS dataset, underscoring its validity and usefulness for research purposes \cite{li-2020, xu-2022, nabian-2019}. The similarities between the current dataset and the Railroad Competition dataset include historical train performance and infrastructure data. However, the Railroad Competition dataset is limited to four months in 2017, whereas the current dataset spans the past 5.5 years, providing a more extensive and recent data source. Additionally, the Railroad Competition studies focus on predicting delays in minutes at a station within a 20-minute timeframe, which is not a binary classification problem but a regression task. In contrast, the current research treats delay prediction as a binary classification problem, suitable for the specific objective of testing the generalizability of the models proposed by Lei et al \cite{lei-2022} on the Dutch railway. A preview of the original NS dataset can be found in Table \ref{tab:preview_dataset} in the Appendix.

\subsubsection{Transformation} \label{sec:data_trans} 
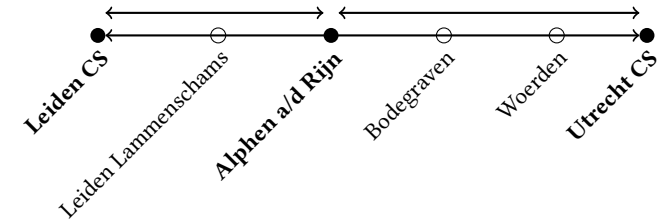
\begin{figure}[htbt]
\centering
\begin{tikzpicture}
    \node at (-0.1,0) [circle,fill,inner sep=2pt] {};
    \node[below left, rotate=45] at (-0.1,0) {\textbf{Leiden CS}};
    \node at (1.5,0) [circle,draw,inner sep=2pt] {};
    \node[below left, rotate=45] at (1.5,0) {Leiden Lammenschams};
    \node at (3,0) [circle,fill,inner sep=2pt] {};
    \node[below left, rotate=45] at (3,0) {\textbf{Alphen a/d Rijn}};
    \node at (4.5,0) [circle,draw,inner sep=2pt] {};
    \node[below left, rotate=45] at (4.5,0) {Bodegraven};
    \node at (6,0) [circle,draw,inner sep=2pt] {};
    \node[below left, rotate=45] at (6,0) {Woerden};
    \node at (7.2,0) [circle,fill,inner sep=2pt] {};
    \node[below left, rotate=45] at (7.2,0) {\textbf{Utrecht CS}};
    \draw[thick, <->] (0, 0.3) -- (2.9, 0.3);
    \draw[thick, <->] (3.1, 0.3) -- (7.1, 0.3);
    \draw[thick, <->] (0, 0) -- (7.1, 0);
\end{tikzpicture}
\caption{Example of three trajectories: Leiden Centraal-Utrecht Centraal, Leiden Centraal-Alphen a/d Rijn, and Alphen a/d Rijn-Utrecht.}
\label{fig:example_trajectory}
\end{figure}

The NS dataset underwent a series of transformations to facilitate both exploratory analysis and model construction. Each row in the original data represents a stop at a station, with each service including at least a departure and an arrival at a station (i.e., two rows). For each stop, the dataset provides the station name, arrival and departure times, delays, and cancellations. The focus of this research is on services rather than individual rides due to the significant difference in their numbers and the nature of delays. Specifically, there are approximately ten times more rides than services, with many involving stops in close proximity (e.g., Amsterdam Zuid and Amsterdam Rai). Delays at such closely spaced stops are often not indicative of broader delay patterns; hence, they are aggregated into services to provide a more meaningful analysis of delay trends within the network. These services are defined as trajectories, referring to the source and target stations of a train ride. The efforts to fund trajectory optimization studies substantiate the use of trajectories in the research field \cite{kouzoupis-2023}. An example of such trajectories can be found in Figure \ref{fig:example_trajectory}, where three trajectories \cite{NS-trajecten} are displayed. 

A critical metric calculated during the data transformation process is the final arrival delay. This metric, representing the delay at the last stop of each trajectory, is chosen because it directly impacts passengers' travel plans and serves as a measure of service reliability from an operational perspective \cite{mou-2019, shi-2021, zhang-2022}. The final arrival delay provides a clear and stable target for predictive modeling, ensuring insights for improving the railway system's efficiency. Other potential metrics, such as maximum delay, mean delay, or departure delays, were considered but ultimately not selected. These alternatives do not capture the overall punctuality and introduce variability that may not accurately reflect the end-to-end travel experience. Consequently, the focus on the final arrival delay ensures a robust and consistent measure for the modeling purposes of trajectories.

To maintain the monthly timeframe, aligning with the baseline study \cite{lei-2022}, the dataset had to be further transformed. Grouping trajectories by month allowed for the evaluation of topological features on trajectories. However, this grouping introduced the challenge of handling multiple counted trains on a trajectory, which results in a non-binary environment (e.g., 5 out of 10 trains on a trajectory had a final arrival delay). To address this, a variable for the proportion of trains with delays on each trajectory was created. After that, a threshold was established to determine significant delays, ensuring the classification remained binary. The proportion of delayed trains was calculated with the following formula:
\[\text{proportion delayed} = \frac{\text{nr. of trajectories}}{\text{nr. of trajectories with final arrival delay}}\]\\
The 50th percentile was then used as a threshold to classify a trajectory as significantly delayed. This method provides a balanced distribution of delayed and non-delayed trajectories, making the dataset suitable for machine learning models and ensuring robustness for training and evaluation \cite{japkowicz-2002, he-2009}. Percentiles, including the 50th percentile, are commonly used in various fields to define anomalies or significant events \cite{narayanan-2011, kim-2018}. This method is based on the distribution of data and helps in setting a natural cutoff point \cite{chandola-2009}. The use of the 50th percentile means that over all the years, 50\% of the delayed edges are classified as significantly delayed, and the other 50\% as not significantly delayed. This calculation with the median sets the threshold for a significantly delayed edge at 21\%, meaning that when a unique edge is delayed more than 21\% of the time, it is classified as significantly delayed.

\subsubsection{Data Cleaning} Data cleaning was performed after the data transformation process. For the US air dataset, data cleaning did not involve many steps. Most importantly, rows with identical source and target cities (indicating no flight) and rows with zero weight (indicating no flight) were removed after grouping the flights by year and month. This clean and transformed dataset can be reviewed in Table \ref{tab:flight_data2} in the Appendix. 

For the transformed dataset of the Dutch railway network, the focus is on train rides provided by NS to make the research specific and because it is the leading provider of train rides in the Netherlands. All other data from providers like Arriva, Conexxion, and QBuzz were removed. Furthermore, NS has by far the most data (almost 70\%), showing its important role in the Dutch railway network. Services by the NS that concern replacement buses or taxis are also removed from the dataset because the focus is on the railway network. In the original dataset, there are no NaN values to be deleted or outliers to handle. There are some NaN values present in the dataset but those have an important meaning, for example, if the arrival delay column is empty, this means that no arrival was planned. These kinds of data attributes were handled during the data transformation step. Handling NaN values in the dataset was crucial for ensuring accurate analysis. NaN values in the 'Arrival Delay of Last Stop' column were set to -1 to indicate unfinished trajectories, while NaN values in the 'Last Arrival Cancelled' column were set to True to signify cancellations. During data aggregation, NaN values in columns such as Final arrival delay and Intermediate arrival delays were filled with 0 to ensure consistency. This careful treatment of NaN values allowed for accurate calculations of the 'Proportion Delayed', which should not take completely canceled rides into consideration.

It is important to note that the dataset used is an archive and not a planned timeline. Furthermore, only trajectories with their source and target station inside the Netherlands were analyzed. This research aimed to analyze the topological features specific to the Dutch railway network. Including trajectories from other countries could introduce variations in network topology that might not be relevant or could confuse the analysis. Lastly, a threshold was established for the minimum number of rides counted per month to ensure data relevance and reliability. Many trajectories were recorded only once or twice a month, which likely indicates substitute transport or non-recurring routes. Including such infrequent data in the model would be impractical, as delays or the absence thereof on these routes could undermine the overall network performance. Therefore, a threshold of at least four rides per month was set, meaning that only routes with a minimum of one weekly train ride are included in the model. By establishing minimum frequency thresholds, routes with sporadic services that do not provide meaningful insight into the overall performance of the transportation network can be excluded \cite{otto-2018}. The final NS dataset used for analysis and model implementation can be found in Table \ref{tab:final_dataset} in the Appendix and an overview of the data transformations and their descriptions is in Table \ref{tab:data-overview}.

\subsubsection{Exploratory Data Analysis} To compare the distributions in both datasets, the number of data points per month (unique edges) and the proportion of those that are True for the dependent variable (e.g., removed or significantly delayed) were evaluated. For the US air data, the number of unique edges per month ranges between 6000 and 7000, representing unique flight connections. Of these unique edges, the proportion that is removed in the subsequent month is between 20\% and 30\% as shown in Figure \ref{fig:us_data} in the Appendix. Both of these counts are relatively stable, except for a downward peak when COVID-19 started in 2020. The proportion of edges that have been removed over the entire dataset's timeframe is 32.7\%, showing an imbalanced dataset.

For the NS data, the number of unique edges per month ranges between 350 and 550, and the proportion of significantly delayed edges varies between 0.1 and 0.9. These counts are relatively unstable, as shown in Figure \ref{fig:ns_data} in the Appendix. Furthermore, the fraction of significantly delayed edges is calculated based on the 50th percentile, as described in section \ref{sec:data_trans} so over all years, 50\% of the trajectories are labeled as significantly delayed. It is noticeable that the number of data points and the movement of the data are different in both cases. Furthermore, the increase in delayed links is visible for the NS data, signifying the relevance of this research.

While the initial phase of data exploration involved analyzing the number of trajectories counted per month for both datasets, it is also valuable to see the new NS network as a whole to gain a comprehensive understanding. Flights and trajectories are defined as connections grouped by source-target pairs. In the case of trajectories, this approach emphasizes network-wide patterns instead of isolated stopovers, while in the case of US flights, this decision was already made \cite{lei-2022}. In Figure \ref{fig:map} in Appendix \ref{sec:map}, the trajectories and proportion of delays in the network for April 2024 are displayed.

\subsection{Model Construction} 
The baseline work focuses on predicting removed links in fast-changing networks by leveraging a variety of machine-learning techniques and network topology features \cite{lei-2022}. This approach is implemented for both the US air data to validate the work and identify its limitations as well as for the NS data to evaluate the generalizability of the baseline model and assess its effectiveness in predicting delayed trajectories in the Dutch railway network. Additionally, the model is improved by incorporating new classifiers to explore the possibility of improved model performance and robustness by testing new classifiers. 

\subsubsection{Feature Extraction}
A comprehensive set of features was derived from both datasets for the prediction models. The primary model framework used by Lei et al. includes several key steps. After the data is cleaned and grouped by month, relevant features are extracted. For each time period, a graph is constructed where nodes represent airports and edges represent connections (flights), just like in Figure \ref{fig:map} but then for the US air network. These graphs of consecutive months are then compared to create the dependent variable, which is whether a link has been removed (e.g., the link is present in month x but not in month x + 1). For all graphs, the weight (number of flights) is provided to capture the strength of connections. Then, various topological features are calculated for each edge in the graph. These features include both weighted and unweighted variants of Common Neighbors, Resource Allocation, Preferential Attachment, Jaccard coefficient, Adamic-Adar index, Salton index, Sorensen index, Hub Promoted index, Hub Depressed index, Leicht-Holme-Newman index, and Local Path index. An overview of the weighted and unweighted topological features can be found in Appendix \ref{sec:featuresets}. Eventually, four different models are considered in the baseline work using unweighted topological features, weighted topological features, edge weights, and unweighted topological features + edge weights as feature vectors for the best-performing classifier.

In the NS data, graphs are also created for each month; only in this way, it does not make sense to compare the graphs of consecutive months to look for the removed edges because the dependent variable, 'Significant Delay', is already present in the data. Nevertheless, the same topological features will be calculated for each graph.

A limitation in the baseline work is the small number of topological features that are tested \cite{lei-2022}. While the Edge Betweenness Centrality and Edge Current Flow Betweenness Centrality were additionally evaluated, these features did not improve the model. A potential enhancement to the baseline model is the inclusion of Node Centrality Measures (NCM). It is argued that if critical nodes can be identified, they can expose the significance of their elements in transportation networks \cite{stamos-2023}. The motivation for including the node centrality measures is three-fold. Firstly, it is obvious that the edge weights cannot be used as a feature vector for the NS data since the dependent variable is counted using these edge weights and this would cause data leakage \cite{kaufman-2012, nassar-2023}. Secondly, centrality measures are a powerful tool in network theory that can be used to understand the importance of nodes in a network \cite{stamos-2023}, which can be a good addition for predicting removed links in the US air network. Lastly, the unique characteristics of the Dutch railway network, particularly the delayed links, benefit from this enhancement. By incorporating centrality measures, the analysis gains valuable insights into the significance and influence of individual stations within the network, thereby providing a deeper understanding of the network's dynamics and potential points of delays. Specifically, the following centrality measures were added: Degree Centrality, which indicates the number of direct connections a node has; Closeness Centrality, which reflects how close a node is to all other nodes in the network based on the shortest paths; and Node Strength, which represents the sum of weights of all edges connected to a node, indicating the total traffic or activity through the node. The definitions of the measures can be found in Table \ref{tab:ncm}. The addition of these centrality measures enhances the model's ability to capture the importance of stations in the network. This addition aligns with the objective of leveraging machine learning techniques to predict delays, considering the significance of network features.

\begin{table}[htbt]
    \centering
    \begin{tabular}{p{1.5cm} p{2cm} p{3.9cm}}
        \toprule
        \textbf{Feature} & \textbf{Definition} & \textbf{Description} \\
        \midrule
        Degree Centrality & $k_i = \sum_{j=1}^{N} A_{ij}$ & Degree centrality of a node \(i\) is the sum of the elements of the adjacency matrix \(A\) at position \(i, j\), where \(N\) is the number of nodes. \\
        Closeness Centrality & $Cl_i = \frac{1}{\sum_{j} d(i, j)}$ & Closeness centrality of a node \(i\) is the reciprocal of the sum of the shortest path distances \(d(i, j)\) from node \(i\) to all other nodes \(j\) in the graph. \\
        Node Strength & $s_i = \sum_{j=1}^{N} w_{ij}$ & Node strength of a node \(i\) is the sum of the weights \(w_{ij}\) of the links connecting node \(i\) to all other nodes \(j\). \\
        \bottomrule
    \end{tabular}
    \caption{The definitions and descriptions of the node centrality measures (NCM). These are calculated for the source and target nodes.}
    \label{tab:ncm}
\end{table}

In conclusion, the feature sets that will be evaluated on the models for both datasets are the basic Topological Features (TF), the Weighted Topological Features (WTF), and the Node Centrality Measures (NCM).
\subsubsection{Model Implementation} \label{sec:model_implementation} After all features have been extracted for both networks, the focus will be on model training and testing. In the baseline work, 27 of the most common supervised classification algorithms \cite{pedregosa-2011, chen-2016} are tested on the balanced training set to determine the best performer based on balanced accuracy, F1 score, and ROC AUC. Among the algorithms tested, XGBoost is identified as the best-performing classifier with the lowest variance among high-performing algorithms in terms of balanced accuracy. The best hyperparameters for the XGBoost model are already defined in the baseline code and can be reviewed in Appendix \ref{sec:xgb_params}. These hyperparameters will be used in the implementation of US air data as well as NS data using the XGBoost Classifier to ensure similar circumstances.

Due to the imbalance in the US air dataset, the RandomUnderSampler from the imblearn library is employed. This technique balances the dataset by randomly undersampling the majority class to ensure an equal representation of both classes during model training, which helps mitigate bias in the model predictions.

Model validation is performed using two primary methods: simultaneous and non-simultaneous testing. Simultaneous testing involves training and testing the model on data from the same period. This approach ensures that the model is evaluated within the same temporal context. Non-simultaneous testing, on the other hand, involves training the model on data from one period and testing it on data from subsequent periods to evaluate the model's ability to generalize over time. 

Feature importance analysis is conducted using SHAP values \cite{lundberg-2017, lundberg-2018}. SHAP values are used to interpret the importance of features in the XGBoost model, providing a way to explain the output of the model by attributing the contribution of each feature to the prediction \cite{marcilio-2020}. The SHAP function computes these values, allowing for a detailed understanding of which features most influence the model's predictions. By implementing this comprehensive framework, the results of the prediction can be evaluated. The combination of simultaneous and non-simultaneous testing ensures both immediate and long-term generalizability of the model, while SHAP values provide transparency into feature importance, enhancing the interpretability of the model's predictions.

\begin{figure}[H]
\includegraphics[width=8cm]{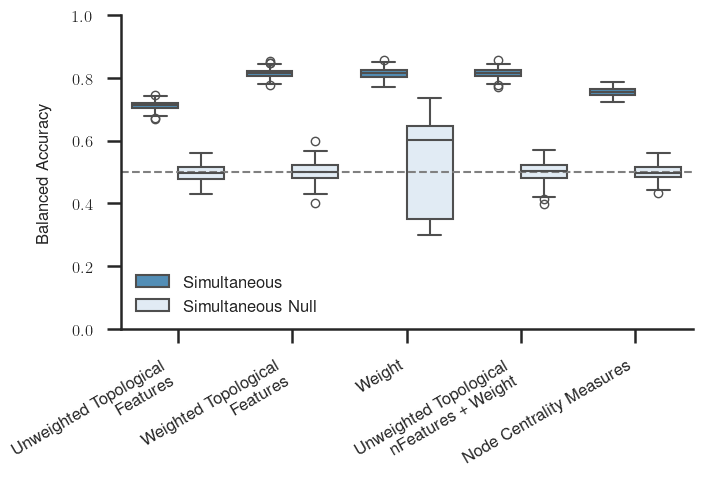}
\centering
\caption{The balanced accuracy boxplots for different feature sets during simultaneous testing on US air data.}
\label{fig:us_met_sim}
\end{figure}

\begin{figure}[H]
\includegraphics[width=8cm]{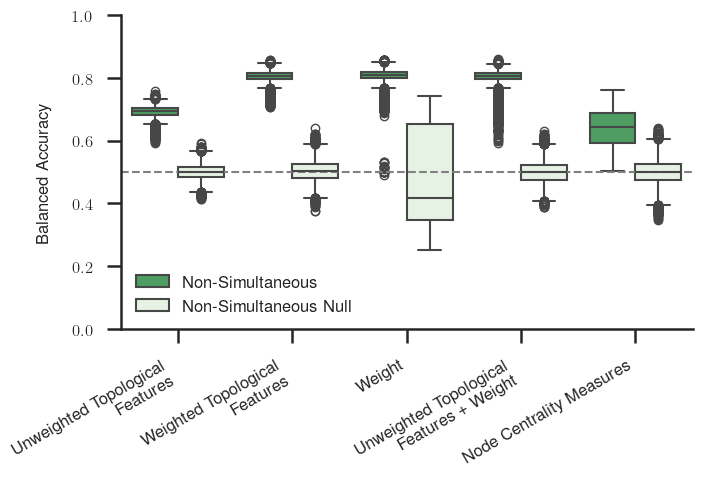}
\centering
\caption{The balanced accuracy boxplots for different feature sets during non-simultaneous testing on US air data.}
\label{fig:us_met_nonsim}
\end{figure}

The performance of the implemented baseline model on US air data is evaluated using balanced accuracy against the null-predictions. The results are presented in Figures \ref{fig:us_met_sim} and \ref{fig:us_met_nonsim}, demonstrating the implementation’s similarity to Lei et al.'s work \cite{lei-2022}.

\subsubsection{Improvements} In addition to incorporating node centrality measures, this research explores the effectiveness of several other classifiers beyond XGBoost on the NS dataset. The aim of this research, besides testing the generalizability of Lei's model, is to be able to predict delayed trajectories as accurately as possible. The new classifiers tested include Random Forest and Decision Tree, next to Gradient Boosting, Ada Boost, and Logistic Regression, which were the best-performing classifiers for the US air data \cite{lei-2022}. Adding these two classifiers broadens the scope of the investigation to determine if performance can be further enhanced with alternative machine learning algorithms. The reason that these two algorithms were chosen is that, next to logistic regression, they are commonly used to perform binary classification tasks \cite{vandebijl2022dutch} with overall good results. Besides that, Random Forest \cite{nabian-2019} as well as Decision Trees \cite{li-2020} have been used in previous works on delay prediction.

Furthermore, hyperparameter optimization was conducted using RandomizedSearchCV with 10-fold cross-validation. Hyperparameter optimization is a crucial step in enhancing the performance of machine learning models. Traditional methods such as grid search, which exhaustively searches over specified parameter values, are computationally expensive and inefficient, especially with high-dimensional parameter spaces. RandomizedSearchCV addresses these limitations by sampling a fixed number of hyperparameter combinations from specified distributions, thus providing a more efficient and often equally effective method for hyperparameter tuning \cite{bergstra-2012}. The decision to use 10-fold cross-validation ensures that the model is evaluated on different subsets of data, providing a more robust estimate of its performance and reducing the risk of overfitting. It is chosen over the default 5-fold cross-validation to explore a wider context and because it demonstrated better performance in previous studies \cite{belete-2021, malakouti-2023}. The specific hyperparameter grids used for each classifier in this research are detailed in Table \ref{tab:params} in the Appendix. Each grid is created to capture the most relevant parameters for the respective models, facilitating an effective and comprehensive search for optimal configurations.

All classifiers were subject to the same evaluation process, involving simultaneous and non-simultaneous testing. This approach ensures a fair comparison of their performance across different feature sets (TF, WTF, NCM). The results of these tests were evaluated based on balanced accuracy, F1 score, and ROC AUC, providing a comprehensive assessment of their effectiveness.

\begin{figure}[htbt]
    \centering
    \begin{tikzpicture}[
        ovalnode/.style={ellipse, draw=black, thick, minimum height=1cm, minimum width=1.8cm, align=center, text width=2.5cm, inner sep=0.5mm, font=\small},
        rectnode/.style={rectangle, draw=black, thick, minimum height=1cm, align=left, text width=3.7cm, inner sep=1mm, font=\small},
        dashedrect/.style={rectangle, draw=black, thick, dashed, minimum height=1cm, align=center, text width=1cm, inner sep=1mm, font=\small},
        font=\footnotesize]
    \node[ovalnode] (data) {Transformation\\(NS Dataset)};
    \node[ovalnode, below=1cm of data] (feature) {Feature Extraction};
    \node[ovalnode, below=1.3cm of feature] (classification) {Classification};
    \node[ovalnode, below=1.1cm of classification] (evaluation) {Evaluation};
    \node[ovalnode, below=0.8cm of evaluation] (longterm) {Long Term\\Prediction};
    \draw[->] (data) -- (feature);
    \draw[->] (feature) -- (classification);
    \draw[->] (classification) -- (evaluation);
    \draw[->] (evaluation) -- (longterm);
    \node[rectnode, right=0.7cm of data, anchor=west] (desc1) {
        - \textbf{Extract trajectorie}s \\
        - \textbf{Calculate proportion delayed} \\
        - \textbf{Determine significant delay (50th percentile)}
    };
    \node[rectnode, right=0.7cm of feature, anchor=west] (desc2) {
        - Unweighted TF \\
        - Weighted TF \\
        - \st{Weight} \\
        - \st{Unweighted TF + Weight} \\
        - \textbf{Node Centrality Measures}
    };
    \node[rectnode, right=0.7cm of classification, anchor=west] (desc3) {
        - Most used supervised classification algorithms \\
        - Simultaneous vs. non-simultaneous testing \\
        - Performance per feature sets \\
        - \textbf{New classifiers + RandomizedSearchCV} 
    };
    \node[rectnode, right=0.7cm of evaluation, anchor=west] (desc4) {
        - Balanced Accuracy \\
        - SHAP Values \\
        - \textbf{Compare performances on different datasets}
    };
    \draw[->] (data.east) -- (desc1.west);
    \draw[->] (feature.east) -- (desc2.west);
    \draw[->] (classification.east) -- (desc3.west);
    \draw[->] (evaluation.east) -- (desc4.west);
    \draw[dashed, thick] ([xshift=-0.8cm, yshift=0.5cm]data.north west) -| ([xshift=0.8cm, yshift=0.5cm]data.north east) -| ([xshift=0.8cm, yshift=-0.5cm]evaluation.south east) |- ([xshift=-0.8cm, yshift=-0.5cm]evaluation.south west) |- cycle;
    \draw[thick] ([xshift=-0.6cm, yshift=0.5cm]feature.north west) -| ([xshift=0.6cm, yshift=0.5cm]feature.north east) -| ([xshift=0.6cm, yshift=-0.4cm]longterm.south east) |- ([xshift=-0.6cm, yshift=-0.4cm]longterm.south west) |- cycle;
    \end{tikzpicture}
    \caption{Project flow highlighting the research scope and methodologies.}
    \label{fig:flowchart}
\end{figure}
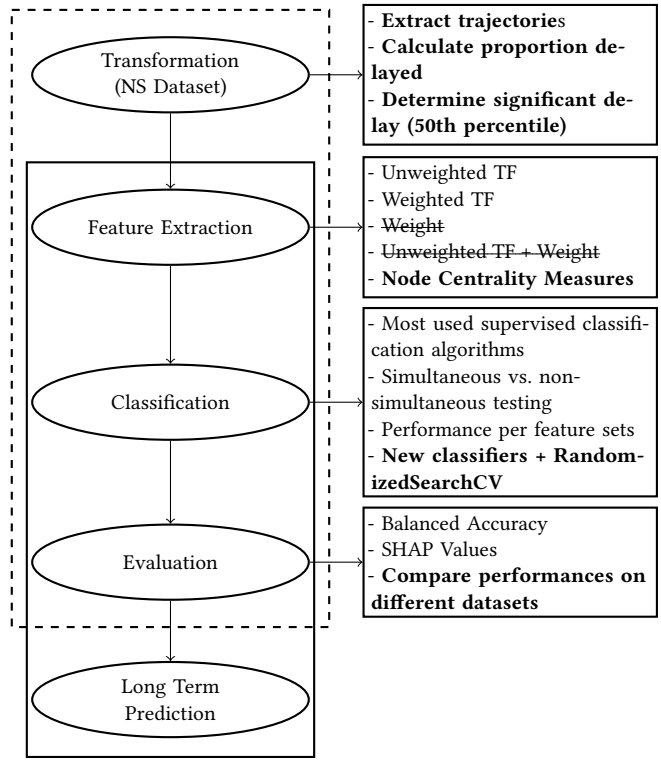

A visualization of the project flow and improvements to the baseline model can be seen in Figure \ref{fig:flowchart}. The components within the dashed line represent elements specific to this research, while the components within the continuous line represent the work by Lei et al. Additionally, bold text highlights additions made in this research compared to the baseline, and strikethrough text indicates elements from the baseline that were not considered. 

\subsection{Evaluation \& Validation} The performance of the improved models trained on NS data, including the centrality measures as a feature set, is compared against the baseline models used by Lei et al. This comparison is based on balanced accuracy to ensure robustness when comparing the two datasets and to prevent bias in the classifier \cite{brodersen-2010}. This metric avoids bias by providing a symmetric evaluation of performance across all classes, ensuring that the results are not distorted by class imbalances. Balanced accuracy is also the metric used for the evaluation of the performance of the added algorithms, next to F1 scores and the ROC AUC. On the one hand, these metrics were chosen because they are also used in baseline work, on the other hand; the F1 score is useful in situations where the costs of false positives and false negatives have to be balanced \cite{m-2015} and the ROC AUC is in general a good metric to measure a model's ability to distinguish between positive and negative classes \cite{boyd-2013, naidu-2023}. 

The implemented models undergo both simultaneous and non-simultaneous testing. The data is split into training and testing sets using a time-based split function. This function ensures that each month has data in both the training and testing sets, maintaining the temporal integrity of the data. Specifically, 70\% of the data from each period is randomly sampled into the training set, and the remaining 30\% into the testing set \cite{lei-2022}. This approach mitigates potential temporal leakage and ensures that the model’s performance is not wrongly based on future information in the training set.

The baseline model's performance on the US air data was evaluated to validate the accuracy of our implementation. Figures \ref{fig:us_met_sim} and \ref{fig:us_met_nonsim} demonstrate the balanced accuracy of all feature sets during simultaneous and non-simultaneous testing, respectively. The results closely match those reported by Lei et al., validating our implementation and allowing us to apply the model to the NS dataset confidently.
By replicating Lei et al.'s methodology and rigorously applying it to the context of the Dutch railway network, we test the generalizability of their model for delay prediction instead of removed link prediction. This involved implementing their work to validate its correctness, identifying limitations, adding an additional feature set (node centrality measures) to overcome these limitations, and applying the exact same model to the NS dataset. The comparison of model performances on the NS data, particularly in terms of balanced accuracy, provides insights into the effectiveness of the implemented models in predicting delays within a different network. Furthermore, all results will be reflected against their null predictions to establish a baseline within each model, helping to assess performance and detect potential outliers \cite{walters-2021, sun-2022}. This comprehensive evaluation framework ensures that improvements are clearly demonstrated and performance validated against the baseline model.

\section{Results}
\label{sec:results}
In this section, the results of the methodology will be evaluated by following the structure of the research questions. 

\subsection{Implementation \& Improvements Baseline}
The first sub-research question focuses on how the machine learning models and methodologies used by Lei et al. can be implemented and improved for their effectiveness in predicting missing links in fast-changing networks. The models 
used by Lei et al. were successfully implemented using the US air dataset, as is already shown in section \ref{sec:model_implementation}. The results confirm the effectiveness of their approach in predicting removed links when performing simultaneous testing as well as non-simultaneous testing for all feature sets. The centrality measures (degree centrality, closeness centrality, and node strength) that are added as an improvement to their models showed almost similar results as the other feature sets during simultaneous testing on the US air data, as can be seen in Figure \ref{fig:us_met_sim}. For non-simultaneous testing, the centrality measures perform a bit worse than the other feature sets and introduce more variance, as can be seen in Figure \ref{fig:us_met_nonsim}.
The results prove that the models by Lei et al. have good performance for the different feature sets in both testing scenarios. This is an important first step for the validation of the models before applying them to the NS dataset. The confusion matrices \ref{fig:us_cm_sim} and \ref{fig:us_cm_nonsim} in Appendix \ref{sec:cm_us} also show that the centrality measures have performance comparable to the topological features. 

\begin{figure}[H]
\includegraphics[width=8cm]{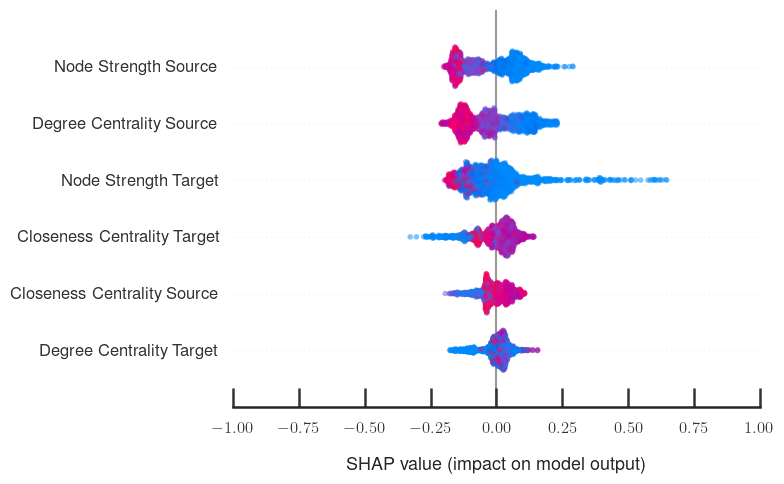}
\centering
\caption{SHAP values for the node centrality measures on US air Data.}
\label{fig:us_shap_ncm}
\end{figure}

\begin{figure}[H]
\includegraphics[width=8cm]{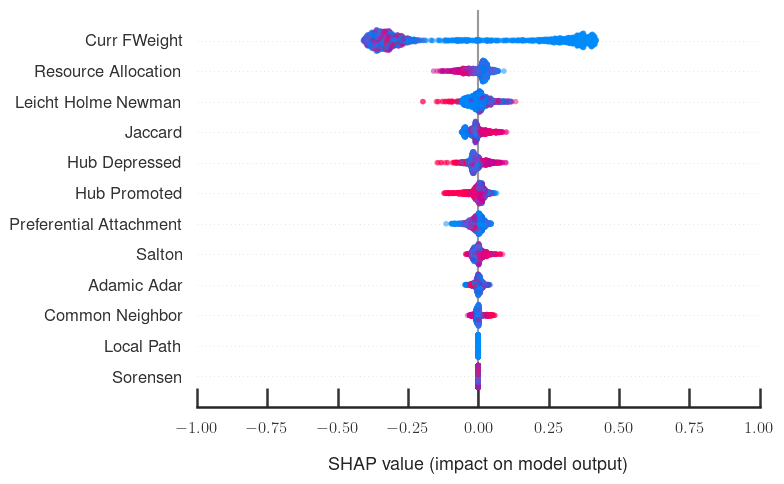}
\centering
\caption{SHAP values for the unweighted topological features + edge weights on US air Data.}
\label{fig:us_shap_btfw}
\end{figure}

Furthermore, when zooming in on the centrality measures and evaluating their importance using SHAP values in Figure \ref{fig:us_shap_ncm}, it is seen that node strength (source station) and degree centrality (source station) are particularly influential in predicting missing links. This indicates that source nodes with higher connectivity and interaction volumes are critical for network stability. Comparing these SHAP values to those of the unweighted topological features in combination with edge weights in Figure \ref{fig:us_shap_btfw}, it is observed that the centrality measures are less consistent in their importance and have less impact in general. Especially when looking at the impact of 'Curr Fweight', the weight of an edge in the current time frame, it is observed that this feature has a strong effect on the output. This is rather obvious since if this weight is high, there is little chance that an edge will be removed in the next month. For this reason, it is a good idea to add the centrality measures as a feature set. In addition, the weights of the connections cannot be used as a feature for the NS data anyway because this value is used to calculate the dependent variable, namely the significant delay. 

\subsection{Implementation NS Data} 
\begin{figure}[H]
\includegraphics[width=8cm]{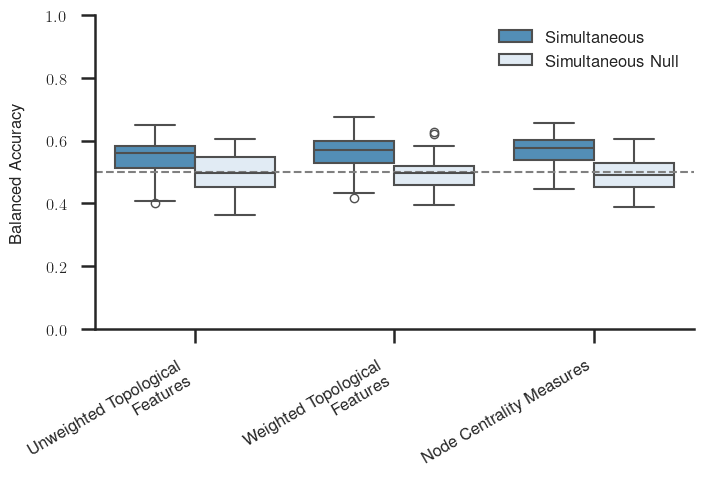}
\centering
\caption{The balanced accuracy boxplots for different feature sets during simultaneous testing on NS data.}
\label{fig:nl_met_sim}
\end{figure}

\begin{figure}[H]
\includegraphics[width=8cm]{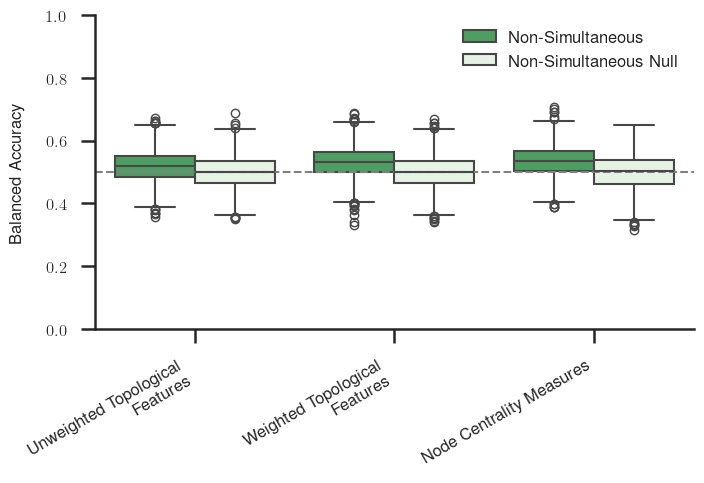}
\centering
\caption{The balanced accuracy boxplots for different feature sets during non-simultaneous testing on NS data.}
\label{fig:nl_met_nonsim}
\end{figure}

The second sub-research question concerns the extent to which the improved methodologies and models can be adapted for predicting delays in the NS network. After the transformation of the data and the calculation of the dependent variable, namely, if a trajectory is significantly delayed (> 21\%, see \ref{sec:data_trans}), it is possible to use the NS dataset in the implemented models. It is important to note that the edge weight and the unweighted topological features + weight will be removed from the model since the edge weight cannot be used as a variable. The node centrality measures will be added instead, next to the existing topological features (weighted and unweighted). When evaluating the obtained results in Figures \ref{fig:nl_met_sim} and \ref{fig:nl_met_nonsim} it is observed that the results are worse for the model when trained on the NS dataset with both simultaneous and non-simultaneous testing. While the balanced accuracies are above the null predictions, they are just above 50\% meaning that they are either not good predictors for the delay of edges or the XGBoost Classifier is not suitable for this prediction.

This is also the result observed in the confusion matrices in Appendix \ref{sec:ns_cm}. The accuracy scores seem to be unstable for this dataset. They predict right nearly as often as wrong. The SHAP figures in Appendix \ref{sec:ns_shap} show the same inconsistency. 

For this reason, other classifiers were evaluated to compare their performances across different feature sets and test if they could result in higher balanced accuracies than the baseline model using the XGBoost Classifier. Only testing one model would not be sufficient to conclude that the baseline model is not generalizable for a new dataset. The results for these added classifiers, as shown in Table \ref{tab:classifier_results} for simultaneous testing and Table \ref{tab:classifier_results_non} for non-simultaneous testing, reveal that centrality measures consistently yielded better results than topological features across various classifiers. However, the overall performance remained poor, particularly in non-simultaneous testing scenarios. Furthermore, it becomes clear that the XGBoost Classifier performs best when simultaneous testing, with RandomForest as a close second. It is noticeable that both during simultaneous testing, and non-simultaneous testing, the results are very close to each other. This also applies to the F1 and ROC AUC scores which can be observed in Appendix \ref{sec:metrics}.
These findings indicate that while the improved models, particularly those incorporating centrality measures, show some promise, they do not seem to be able to predict delayed trajectories in the Dutch railway network well. 

\begin{table}[htbt]
\centering
\begin{tabular}{p{3.7cm}p{1cm}p{1cm}p{1cm}}
\toprule
\textbf{Classifier} & \textbf{TF} & \textbf{WTF} & \textbf{NCM} \\
\midrule
AdaBoost & 0.618 & 0.621 & \textbf{0.635} \\ 
DecisionTree & 0.602 & \textbf{0.604} & 0.591 \\ 
GradientBoosting & 0.617 & 0.620 & \textbf{0.623} \\ 
LogisticRegression & \textbf{0.636} & 0.633 & 0.632 \\ 
RandomForest & 0.620 & 0.626 & \textbf{0.645} \\ 
\textbf{XGBoost} & 0.635 & 0.633 & \textbf{0.646} \\ 
\bottomrule
\end{tabular}
\caption{Balanced accuracy scores of classifiers for every feature set during simultaneous testing.}
\label{tab:classifier_results}
\end{table}

\begin{table}[htbt]
\centering
\begin{tabular}{p{3.7cm}p{1cm}p{1cm}p{1cm}}
\toprule
\textbf{Classifier} & \textbf{TF} & \textbf{WTF} & \textbf{NCM} \\
\midrule
AdaBoost & 0.527 & 0.532 & \textbf{0.538} \\ 
DecisionTree & 0.525 & 0.526 & \textbf{0.534} \\ 
GradientBoosting & 0.527 & 0.533 & \textbf{0.545} \\ 
LogisticRegression & 0.525 & 0.535 & \textbf{0.537} \\ 
\textbf{RandomForest} & 0.529 & 0.535 & \textbf{0.550} \\ 
XGBoost & 0.529 & 0.532 & \textbf{0.538} \\ 
\bottomrule
\end{tabular}
\caption{Balanced accuracy scores of classifiers for every feature set during non-simultaneous testing.}
\label{tab:classifier_results_non}
\end{table}

\subsection{Model Performance and Comparison}
To answer the last sub-research question, the balanced accuracies for the XGBoost classifier are merged for both datasets. While the other classifiers showed similar balanced accuracy scores for the NS dataset, the XGBoost Classifier was used to enable the comparison with the baseline article. In Table \ref{tab:final_results} the performances for the different datasets can be compared and it is clear that the model performs better on the US air data when predicting removed links than on the NS data when predicting significantly delayed links, even when adding the additional centrality measures. 

\begin{table}[htbt]
    \centering
    \begin{tabular}{p{2.5cm}p{2cm}p{0.7cm}p{0.7cm}p{0.7cm}}
        \toprule
        \textbf{Testing Type} & \textbf{Dataset} & \textbf{TF} & \textbf{WTF} & \textbf{NCM} \\
        \midrule
        \multirow{2}{1em}{Simultaneous} & \textbf{US air Data} & 0.71 & \textbf{0.81} & 0.76 \\
        & NS Data & 0.64 & 0.63 & \textbf{0.65} \\
        \midrule
        \multirow{2}{1em}{Non-simultaneous} & \textbf{US air Data} & 0.68 & \textbf{0.80} & 0.68 \\
        & NS Data & 0.53 & 0.53 & \textbf{0.54} \\
        \bottomrule
    \end{tabular}
    \caption{Balanced accuracy scores for the different datasets using the XGBoost Classifier in both testing scenarios.}
    \label{tab:final_results}
\end{table}

\section{Discussion}

This research aimed to implement Lei et al.'s machine learning framework, initially designed for forecasting removed links in fast-changing networks, to predict delayed trajectories within the Dutch railway network. The baseline framework demonstrated high balanced accuracy in predicting removed links in the US air network, achieving scores of 0.71-0.81 in simultaneous testing and 0.68-0.80 in non-simultaneous testing. However, the improvements made to the model to implement the Dutch railway data yielded lower balanced accuracy scores, with simultaneous testing scores ranging from 0.63 to 0.64 and non-simultaneous testing scores between 0.52 and 0.54.

The lower performance on the Dutch railway data suggests that while Lei et al.'s approach is effective for predicting removed links in the US air network, it may not be directly applicable to networks with different dynamics and operational characteristics. The additional centrality measures, although theoretically promising, did not significantly enhance the model's predictive performance either. This observation is supported by the higher variance and lower balanced accuracy in the US context for non-simultaneous testing.

Several studies have highlighted the importance of adapting machine learning models and their features to the specific characteristics of the dataset and network being analyzed. For instance, research on public transportation networks in Europe has shown that incorporating domain-specific features such as scheduling information, passenger load, and infrastructure maintenance schedules can significantly improve prediction accuracy \cite{zhu-2018, besinovic-2020}. The unique aspects of railway networks, such as fixed routes and schedules, contrast with the more dynamic nature of air transportation systems, potentially explaining the poorer performance observed in this research.

Even after incorporating other machine learning models commonly used in binary classification, the accuracy scores remained quite similar to those of the XGBoost Classifier. This suggests that these models might be too simplistic for the complex data, or that the issue lies within the data itself, not leading to performance improvements with new models.

\subsection{Limitations \& Future Work}
The results were not as favorable as anticipated, particularly in non-simultaneous testing scenarios. Several factors can be considered to explain this discrepancy. First, certain topological features may be less predictive because the Dutch railway network has different operational characteristics than the US air network. The railway network in the Netherlands is very dense and therefore very sensitive to small timetable changes. As mentioned by the Head of Digitalization Operations at NS, B. van Zaalen (personal communication, June 4, 2024), the ripple effect is a big problem in the network. A small congestion somewhere on a route can cause many delays elsewhere. There can be many different causes for network congestion or delay, and perhaps topological and network features alone are not enough to predict it. The focus on topological features and centrality measures, while inspired by related work \cite{lei-2022, stamos-2023}, might not adequately capture other critical factors such as maintenance schedules, staffing issues, or external disruptions, which are known to affect railway network performance \cite{li-2020}.

The scalability and generalizability of the adapted models also present challenges. While Lei et al.'s framework demonstrated robustness across different networks during simultaneous testing, its adaptation to the Dutch railway network indicates the need for more context-specific modifications. Additionally, the models' even poorer performance in non-simultaneous testing highlights potential issues with temporal generalizability, suggesting that further refinement is necessary to enhance predictive stability over time. 

Following this, another possible limitation lies in the dataset. There is much less available data per month compared to the US air data (more than 10 times less), which could mean that there is not enough training data per month to find the relevant patterns. A suggestion would be to include other providers or look at all rides instead of the focus on trajectories. Furthermore, conducting a comparative analysis with other transportation networks, particularly those with similar operational characteristics to the US air network, could also provide valuable insights into model adaptability, generalizability, and performance.

Additionally, employing more sophisticated machine learning techniques, such as ensemble methods or deep learning architectures, might enhance predictive accuracy. However, it is noticeable that already various frequently used classifiers have been compared and they all show similar output results. This indicates that the model is not the limiting factor. In any case, more consideration will have to be given to the extreme complexity of the Dutch railway network. Future research should explore integrating operational, environmental, and temporal variables to develop a more holistic predictive model.

By addressing these limitations and exploring these future directions, it is possible to develop a more robust and accurate model for predicting delayed trajectories in the Dutch railway network. This research contributes to the broader understanding of the challenges and opportunities in applying machine learning to transportation networks, highlighting the need for context-specific adaptations. 

\section{Conclusion}
\label{sec:conclusion}
The objective of this research was to evaluate the generalizability of Lei et al.'s machine learning framework, which predicts removed links in the fast-changing US Air network, by applying it to predict delayed trajectories in the Dutch railway network. 
To answer the first sub-research question, the XGBoost Classifier can be implemented to predict removed links in fast-changing transportation networks like the US Air data and can be improved by adding the node centrality measures as a feature set. The results show high balanced accuracies for this prediction. Furthermore, these improved models can be constructed to fit with another dataset, namely the NS dataset, as well. However, poor performance was observed when this dataset was used in the improved baseline model compared to the performance of the US air dataset. Even when other classifiers, like the  Decision Tree and Random Forest, were evaluated it did not seem possible to predict delayed trajectories with the baseline framework.

The research findings highlight the challenges of directly transferring methodologies across different transportation networks with distinct operational dynamics. Significant effort was devoted to transforming the NS dataset to fit the implemented model, but these attempts did not yield the desired results. While the models achieved balanced accuracy scores above the null predictions, the overall performance was not good, indicating the need for more context-specific adaptations and the incorporation of diverse feature sets.

The limitations, especially the exclusion of non-topological factors, suggest areas for future research. Integrating operational, environmental, and temporal variables, alongside more advanced machine learning techniques, could enhance predictive accuracy. 

In conclusion, while this research does not provide a definitive solution for predicting delayed trajectories in the Dutch railway network, it offers valuable insights into the complexities and challenges of transportation network forecasting. Future work should focus on developing more holistic models that account for the multifaceted and complex nature of railway operations, ultimately contributing to more reliable and efficient transportation systems.


\bibliographystyle{plain}
\bibliography{references.bib}

\appendix
\newpage
\onecolumn
\section*{Appendix}

\section{Data Previews}
\begin{table*}[htbt]
\centering
\resizebox{\textwidth}{!}{%
\begin{tabular}{lllllllllllllllll}
\toprule
\textbf{DEP\_SCHEDULED} & \textbf{DEP\_PERFORMED} & \textbf{SEATS} & \textbf{PASSENGERS} & \textbf{DISTANCE} & \textbf{UNIQUE\_CARRIER} & \textbf{UNIQUE\_CARRIER\_NAME} & \textbf{ORIGIN\_AIRPORT\_ID} & \textbf{ORIGIN} & \textbf{ORIGIN\_CITY\_NAME} & \textbf{ORIGIN\_STATE\_ABR} & \textbf{DEST\_AIRPORT\_ID} & \textbf{DEST} & \textbf{DEST\_CITY\_NAME} & \textbf{DEST\_STATE\_ABR} & \textbf{YEAR} & \textbf{MONTH} \\ \midrule
0.00 & 17.00 & 0.00 & 0.00 & 2846.00 & "9S" & Southern Air Inc. & 13930 & ORD & Chicago, IL & IL & 10299 & ANC & Anchorage, AK & AK & 2008 & 10 \\ 
0.00 & 1.00 & 262.00 & 0.00 & 408.00 & "DL" & Delta Air Lines Inc. & 14112 & PIE & St. Petersburg, FL & FL & 10397 & ATL & Atlanta, GA & GA & 2008 & 10 \\ 
0.00 & 1.00 & 68.00 & 24.00 & 321.00 & "09Q" & Swift Air, LLC d/b/a Eastern Air Lines d/b/a Eastern & 12264 & IAD & Washington, DC & VA & 11057 & CLT & Charlotte, NC & NC & 2008 & 11 \\ 
0.00 & 1.00 & 4.00 & 0.00 & 53.00 & "2O" & Island Air Service & 10324 & AOS & Amook Bay, AK & AK & 10170 & ADQ & Kodiak, AK & AK & 2008 & 11 \\ 
0.00 & 1.00 & 0.00 & 0.00 & 1321.00 & "8C" & Air Transport International & 12206 & HRL & Harlingen/San Benito, TX & TX & 15295 & TOL & Toledo, OH & OH & 2008 & 5 \\ 
0.00 & 1.00 & 0.00 & 0.00 & 961.00 & "AMQ" & Ameristar Air Cargo & 12206 & HRL & Harlingen/San Benito, TX & TX & 15016 & STL & St. Louis, MO & MO & 2008 & 5 \\ 
0.00 & 1.00 & 6.00 & 2.00 & 30.00 & "GV" & Grant Aviation & 11336 & DLG & Dillingham, AK & AK & 14037 & PCA & Portage Creek, AK & AK & 2008 & 5 \\ 
0.00 & 1.00 & 3.00 & 1.00 & 107.00 & "KAH" & Kenmore Air Harbor & 11762 & FRD & Friday Harbor, WA & WA & 13865 & OLM & Olympia, WA & WA & 2008 & 5 \\ 
0.00 & 1.00 & 5.00 & 1.00 & 30.00 & "KS" & Peninsula Airways Inc. & 11336 & DLG & Dillingham, AK & AK & 14037 & PCA & Portage Creek, AK & AK & 2008 & 5 \\ 
\bottomrule
\end{tabular}%
}
\caption{A preview of the original US data from \href{https://arch.library.northwestern.edu/concern/datasets/pn89d700k}{the online library} created by Lei et al. \cite{lei-2022}. Each row represents a flight with various attributes such as the number departures scheduled, and performed, number of seats, passengers, distance, carrier information, origin, and destination details.}
\label{tab:flight_data1}
\end{table*}

\begin{table*}[htbt]
\centering
\resizebox{\textwidth}{!}{%
\begin{tabular}{p{3cm}p{3cm}p{5cm}p{5cm}p{4cm}p{4cm}}
\toprule
\textbf{YEAR} & \textbf{MONTH} & \textbf{Source} & \textbf{Target} & \textbf{Passengers} & \textbf{Weight} \\ \midrule
2004 & 1 & aberdeen\_sd & jamestown\_nd & 45.0 & 25.0 \\ 
2004 & 1 & aberdeen\_sd & minneapolis\_mn & 2782.0 & 141.0 \\ 
2004 & 1 & aberdeen\_sd & pierre\_sd & 505.0 & 52.0 \\ 
2004 & 1 & aberdeen\_sd & sioux\_falls\_sd & 0.0 & 35.0 \\ 
2004 & 1 & aberdeen\_sd & watertown\_sd & 139.0 & 29.0 \\ 
2004 & 1 & abilene\_tx & dallasfort\_worth\_tx & 3727.0 & 186.0 \\ 
2004 & 1 & abilene\_tx & elko\_nv & 44.0 & 1.0 \\ 
2004 & 1 & abilene\_tx & houston\_tx & 835.0 & 81.0 \\ 
2004 & 1 & abilene\_tx & lubbock\_tx & 0.0 & 44.0 \\ 
\bottomrule
\end{tabular}%
}
\caption{A preview of the cleaned US data. Each row represents a flight connection with various attributes such as year, month, source, target, passengers, and weight showing the number of departures performed.}
\label{tab:flight_data2}
\end{table*}

\begin{table*}[htbt]
\centering
\resizebox{\textwidth}{!}{%
\begin{tabular}{llllllllllllll}
\toprule
\textbf{RDT-ID} & \textbf{Date} & \textbf{Type} & \textbf{Company} & \textbf{Completely cancelled} & \textbf{Partly cancelled} & \textbf{Maximum delay} & \textbf{Station name} & \textbf{Arrival time} & \textbf{Arrival delay} & \textbf{Arrival cancelled} & \textbf{Departure time} & \textbf{Departure delay} & \textbf{Departure cancelled} \\ \midrule
738804 & 01-01-2019 & Intercity & NS & False & False & 1 & Rotterdam Centraal & NaN & NaN & NaN & 2019-01-01T02:0 & 1.0 & False \\ 
738804 & 01-01-2019 & Intercity & NS & False & False & 0 & Delft & 2019-01-01T01:0 & 1.0 & False & 2019-01-01T02:1 & 0.0 & False \\ 
738804 & 01-01-2019 & Intercity & NS & False & False & 0 & Den Haag HS & 2019-01-01T01:1 & 0.0 & False & 2019-01-01T02:2 & 1.0 & False \\ 
738804 & 01-01-2019 & Intercity & NS & False & False & 0 & Leiden Centraal & 2019-01-01T01:2 & 0.0 & False & 2019-01-01T02:4 & 0.0 & False \\ 
738804 & 01-01-2019 & Intercity & NS & False & False & 0 & Schiphol Airport & 2019-01-01T01:3 & 0.0 & False & 2019-01-01T03:0 & 0.0 & False \\ 
13090236 & 29-02-2024 & Intercity & NS & False & False & 0 & Amsterdam Zuid & 2024-03-01T00:4 & 8.0 & False & 2024-03-01T00:4 & 9.0 & False \\ 
13090236 & 29-02-2024 & Intercity & NS & False & False & 0 & Schiphol Airport & 2024-03-01T00:5 & 9.0 & False & 2024-03-01T00:5 & 8.0 & False \\ 
13090236 & 29-02-2024 & Intercity & NS & False & False & 0 & Leiden Centraal & 2024-03-01T00:5 & 8.0 & False & 2024-03-01T00:7 & 8.0 & False \\ 
13092579 & 29-02-2024 & Extra trein & NS & False & False & 0 & Meppel & NaN & NaN & NaN & NaN & NaN & NaN \\ 
\bottomrule
\end{tabular}%
}
\caption{A preview of the original NS dataset from \href{https://www.rijdendetreinen.nl/en/open-data}{rijdendetreinen.nl}. A unique RDT-ID defines a train service/trajectory and every row defines a stop at a station.}
\label{tab:preview_dataset}
\end{table*}

\begin{table}[H]
\centering
\resizebox{\textwidth}{!}{%
\begin{tabular}{llllllllll}
\toprule
\textbf{YearMonth} & \textbf{source} & \textbf{target} & \textbf{Rides planned} & \textbf{Final arrival delay} & \textbf{Final arrival cancelled} & \textbf{Completely cancelled} & \textbf{Intermediate arrival delays} & \textbf{Proportion delayed} & \textbf{Significant Delay} \\ \midrule
        2019-01 & 's-Hertogenbosch & Arnhem Centraal & 57 & 20 & 4 & 0 & 36 & 0.3509 & True \\ 
2019-01 & 's-Hertogenbosch & Den Haag Centraal & 1106 & 99 & 20 & 5 & 853 & 0.0899 & False \\ 
2019-01 & 's-Hertogenbosch & Deurne & 757 & 247 & 8 & 2 & 503 & 0.3272 & True \\ 
2019-01 & 's-Hertogenbosch & Dordrecht & 100 & 23 & 3 & 0 & 44 & 0.2300 & True \\ 
2019-01 & 's-Hertogenbosch & Eindhoven Centraal & 257 & 81 & 1 & 1 & 138 & 0.3164 & True \\ 
2019-01 & 's-Hertogenbosch & Roosendaal & 34 & 9 & 3 & 1 & 15 & 0.2727 & True \\ 
2019-01 & Bergen op Zoom & Amsterdam Centraal & 22 & 14 & 1 & 0 & 22 & 0.6364 & True \\ 
2019-01 & Bergen op Zoom & Roosendaal & 44 & 2 & 4 & 4 & 2 & 0.0500 & False \\ 
2019-01 & Breda & Amsterdam Centraal & 1257 & 727 & 131 & 70 & 948 & 0.6125 & True \\ 
\bottomrule
\end{tabular}%
}
\caption{A preview of the cleaned and transformed NS dataset displaying the unique trajectories per month, their counts, and the dependent variable (most-right column) being the binary value showing if a trajectory was significantly delayed (50th percentile, > 21\%) within that month.}
\label{tab:final_dataset}
\end{table}

\begin{table}[H]
\resizebox{\textwidth}{!}{%
\begin{tabular}{lll}
    \toprule
    \textbf{Data} & \textbf{Number of rows} & \textbf{Description} \\
    \midrule
    Original & 115.266.904 & Service archive from 2019 till April 2024 \\
    Cleaned & 75.620.094 & Filtered on NS and trains (e.g., no replacement busses or taxis) \\
    Grouped & 7.344.087 & Grouped by RDT-ID (e.g., all trajectory rides) \\
    Daily & 559.668 & Daily Aggregation on source - target (all unique trajectories per day)\\
    Monthly & 28.557 & Monthly aggregation on source - target (all unique trajectories per month); Trajectories outside NL removed; threshold for min. 4 rides \\ 
    \bottomrule
    \end{tabular}
    }
\caption{Overview of the different data transformations on the NS dataset, the number of rows for each dataset, and a description of what the dataset entails or how it has been transformed.}
\label{tab:data-overview}
\end{table} \newpage

\section{Exploratory Data Analysis}
\subsection{Dataset size comparison} \label{sec:eda}
\begin{figure}[H]
    \centering
    \includegraphics[width=16cm]{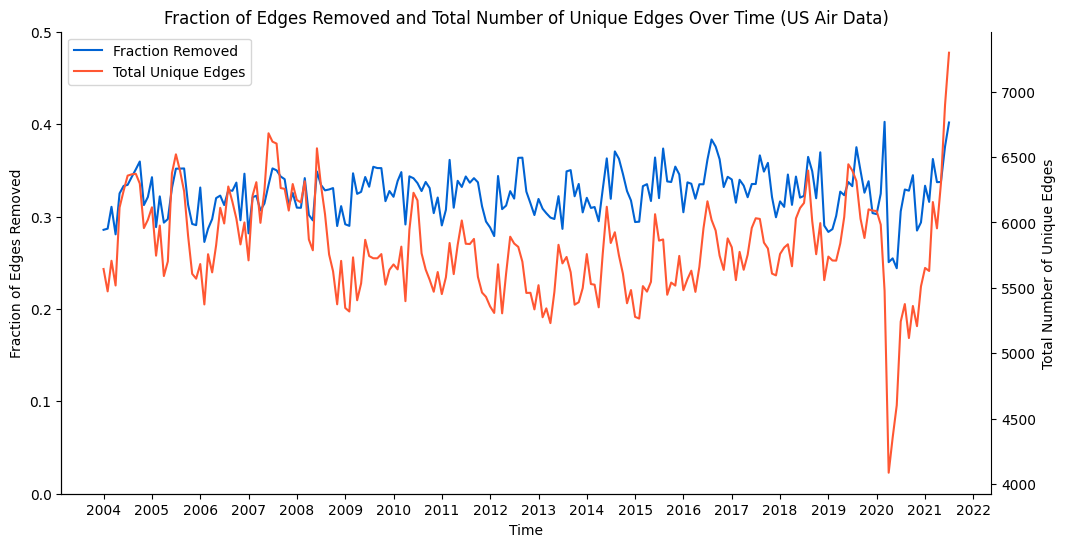}
    \caption{The proportions of removed edges (left axis) and the number of unique edges (right axis) per month for the US air data.}
    \label{fig:us_data}
\end{figure}
    
\begin{figure}[H]
    \centering
    \includegraphics[width=16cm]{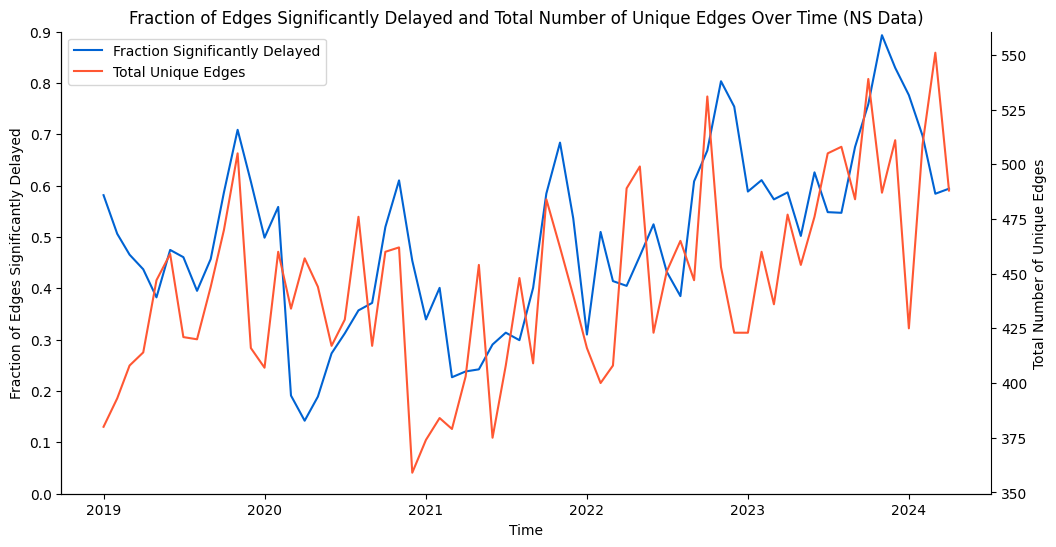}
    \caption{The proportion of significantly delayed edges (left axis) and the number of unique edges (right axis) per month for the NS data.}
    \label{fig:ns_data}
\end{figure} \newpage

\subsection{Map} \label{sec:map}
\begin{figure}[htbt]
\includegraphics[width=16cm]{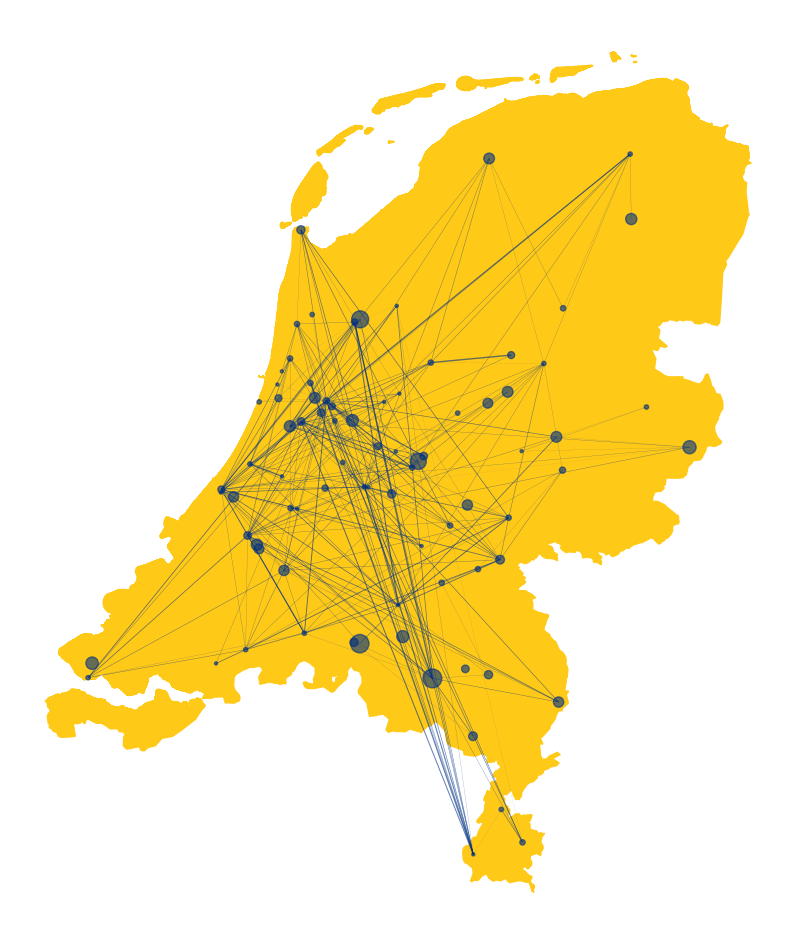}
\centering
\caption{Trajectories in April 2024 with the size of the nodes representing the degree of the nodes in the graph, e.g., the connections (or edges) it has to other nodes. The thickness of the edges displays the proportion of delayed edges.}
\label{fig:map}
\end{figure} 

\newpage

\section{Feature Sets} \label{sec:featuresets}
\subsection{Unweighted Topological Features (TF)}
\begin{table}[H]
\resizebox{\textwidth}{!}{%
    \centering
    \begin{tabular}{>{\raggedright}p{3.5cm} >{\raggedright}p{3cm} >{\raggedright\arraybackslash}p{7.5cm}}
        \toprule
        \textbf{Feature} & \textbf{Definition} & \textbf{Description} \\
        \midrule
        Common Neighbors (CN) & $|\Gamma_i \cap \Gamma_j|$ & The number of common neighbors of nodes $i$ and $j$ \\
        Salton Index (SA) & $\frac{|\Gamma_i \cap \Gamma_j|}{\sqrt{k_i \times k_j}}$ & The number of common neighbors normalized by the geometric average degree of both nodes \\
        Jaccard Index (JA) & $\frac{|\Gamma_i \cap \Gamma_j|}{|\Gamma_i \cup \Gamma_j|}$ & The number of common neighbors normalized by the union of neighbors of both nodes \\
        Sørensen Index (SO) & $\frac{2 |\Gamma_i \cap \Gamma_j|}{k_i + k_j}$ & The number of common neighbors normalized by the average degree of the two nodes \\
        Hub Promoted Index (HPI) & $\frac{|\Gamma_i \cap \Gamma_j|}{\min(k_i, k_j)}$ & The number of common neighbors normalized by the smaller degree of the two nodes \\
        Hub Depressed Index (HDI) & $\frac{|\Gamma_i \cap \Gamma_j|}{\max(k_i, k_j)}$ & The number of common neighbors normalized by the larger degree of the two nodes \\
        Leicht-Holme-Newman Index (LHNI) & $\frac{|\Gamma_i \cap \Gamma_j|}{k_i \times k_j}$ & The number of common neighbors normalized by the product of degrees of the two nodes \\
        Preferential Attachment Index (PA) & $k_i \times k_j$ & The product of the degrees of the two nodes \\
        Adamic-Adar Index (AA) & $\sum_{n \in \Gamma_i \cap \Gamma_j} \frac{1}{\log k_n}$ & The number of common neighbors with each of them normalized by the logarithm of their degree \\
        Resource Allocation Index (RA) & $\sum_{n \in \Gamma_i \cap \Gamma_j} \frac{1}{k_n}$ & The number of common neighbors with each of them normalized by their degree \\
        Local Path Index (LPI) & $S_{ij,2} + \epsilon S_{ij,3}$ & The first term represents the number of paths of length equal to 2 between the node $i$ and $j$. The second term is the number of paths of length equal to 3 between the node $i$ and $j$ damped by parameter $\epsilon$. We set $\epsilon = 0.01$. \\
        \bottomrule
    \end{tabular}
  }
    \caption{The definitions and descriptions of the unweighted topological features (TF).}
    \label{tab:tf}
\end{table}
\newpage

\subsection{Weighted Topological Features (WTF)}
\begin{table}[H]
\resizebox{\textwidth}{!}{%
    \centering
    \begin{tabular}{>{\raggedright}p{3.5cm} >{\raggedright}p{3cm} >{\raggedright\arraybackslash}p{7.5cm}}
        \toprule
        \textbf{Feature} & \textbf{Definition} & \textbf{Description} \\
        \midrule
        Weighted Common Neighbors (WCN) & $\sum_{w_n \in \Gamma_i \cap \Gamma_j} \min(w_{in}, w_{jn})$ & The number of common neighbors weighted by the minimum weight of their connections \\
        Weighted Salton Index (WSA) & $\frac{\text{WCN}}{\sqrt{w_i \times w_j}}$ & The weighted common neighbors normalized by the geometric average weight of both nodes \\
        Weighted Jaccard Index (WJA) & $\frac{\text{WCN}}{w_i + w_j - \text{WCN}}$ & The weighted common neighbors normalized by the union of neighbors of both nodes \\
        Weighted Sørensen Index (WSO) & $\frac{2 \times \text{WCN}}{w_i + w_j}$ & The weighted common neighbors normalized by the average weight of the two nodes \\
        Weighted Hub Promoted Index (WHPI) & $\frac{\text{WCN}}{\min(w_i, w_j)}$ & The weighted common neighbors normalized by the smaller weight of the two nodes \\
        Weighted Hub Depressed Index (WHDI) & $\frac{\text{WCN}}{\max(w_i, w_j)}$ & The weighted common neighbors normalized by the larger weight of the two nodes \\
        Weighted Leicht-Holme-Newman Index (WLHNI) & $\frac{\text{WCN}}{w_i \times w_j}$ & The weighted common neighbors normalized by the product of weights of the two nodes \\
        Weighted Preferential Attachment Index (WPA) & $w_i \times w_j$ & The product of the weights of the two nodes \\
        Weighted Adamic-Adar Index (WAA) & $\sum_{w_n \in \Gamma_i \cap \Gamma_j} \frac{1}{\log w_n}$ & The weighted common neighbors with each of them normalized by the logarithm of their weight \\
        Weighted Resource Allocation Index (WRA) & $\sum_{w_n \in \Gamma_i \cap \Gamma_j} \frac{1}{w_n}$ & The weighted common neighbors with each of them normalized by their weight \\
        Weighted Local Path Index (WLPI) & $W_{ij,2} + \epsilon W_{ij,3}$ & The first term represents the number of paths of length equal to 2 between the node $i$ and $j$. The second term is the number of paths of length equal to 3 between the node $i$ and $j$ damped by parameter $\epsilon$. We set $\epsilon = 0.01$. \\
        \bottomrule
    \end{tabular}
    }
    \caption{The definitions and descriptions of the weighted topological features (WTF).}
    \label{tab:wtf}
\end{table} \newpage

\section{Hyperparameters used for XGBoost Classifier} \label{sec:xgb_params}
\begin{table}[H]
\centering
\begin{tabular}{>{\raggedright}p{3cm} >{\raggedright}p{4cm} >{\raggedright\arraybackslash}p{9cm}}
\toprule
\textbf{Hyperparameter} & \textbf{Value} & \textbf{Explanation} \\
\midrule
lambda & 0.5650701862593042 & L2 regularization term on weights. Increasing this value will make model more conservative. Range: \([0, \infty]\) \\ 
alpha & 0.0016650896783581535 & L1 regularization term on weights. Increasing this value will make model more conservative. Range: \([0, \infty]\)  \\ 
colsample\_bytree & 1.0 & The subsample ratio of columns when constructing each tree. Subsampling occurs once for every tree constructed. \\
subsample & 0.5 & Denotes the fraction of observations to be random samples for each tree. Typically set to 0.5. \\ 
learning\_rate & 0.009 & Step size shrinkage used in update to prevent overfitting. After each boosting step, we can directly get the weights of new features, and \(learning_rate\) shrinks the feature weights to make the boosting process more conservative. Range: \([0, 1]\) \\ 
n\_estimators & 625 & Number of boosting stages to be run. More stages can improve accuracy but may lead to overfitting and higher computation time.\\ 
objective & reg:squarederror & Regression with squared loss.\\ 
max\_depth & 5 & Maximum depth of a tree. Increasing this value will make the model more complex and more likely to overfit. 0 indicates no limit on depth. Range: \([0, \infty]\)\\ 
min\_child\_weight & 6 & Minimum sum of instance weight (hessian) needed in a child. Range: \([0, \infty]\) \\
\bottomrule
\end{tabular}
\caption{Hyperparameters provided in baseline \cite{lei-2022} for the XGBoost Classifier. Explanations are from the official \href{https://xgboost.readthedocs.io/en/latest/parameter.html\#general-parameters}{XGBoost Documentation Website}.}
\label{tab:xgboost_hyperparameters}

\end{table} 
\newpage

\section{Hyperparameter grids for RadomizedSearchCV}
\begin{table}[h!]
\centering
\label{tab:hyperparameters}
\begin{tabular}{l|llp{8cm}}
\toprule
\textbf{Classifier}              & \textbf{Hyperparameter}       & \textbf{Values}                               & \textbf{Explanation} \\ \midrule
\multirow{6}{*}{GradientBoosting} & n\_estimators            & [50, 100, 200, 300]                           & Number of boosting stages to be run. More stages can improve accuracy but may lead to overfitting and higher computation time. \\ \cline{2-4} 
                                 & learning\_rate           & [0.01, 0.05, 0.1, 0.2]                        & Shrinks the contribution of each tree. Lower values require more trees to achieve the same training error, balancing between speed and accuracy. \\ \cline{2-4} 
                                 & max\_depth               & [3, 5, 7, 9]                                  & Maximum depth of the individual trees. Deeper trees can capture more patterns but are more prone to overfitting. \\ \cline{2-4} 
                                 & min\_samples\_split      & [2, 5, 10]                                    & Minimum number of samples required to split an internal node. Prevents overfitting by ensuring splits are based on a sufficient number of samples. \\ \cline{2-4} 
                                 & min\_samples\_leaf       & [1, 2, 4]                                     & Minimum number of samples required to be at a leaf node. Prevents overfitting by ensuring leaves have enough samples. \\ \cline{2-4} 
                                 & max\_features            & ['sqrt', 'log2']                              & Number of features to consider when looking for the best split. Helps in reducing overfitting. \\ \hline
\multirow{2}{*}{AdaBoost}         & n\_estimators            & [50, 100, 200, 400]                           & Number of boosting stages to be run. More stages can improve performance but also increase computation time. \\ \cline{2-4} 
                                 & learning\_rate           & [0.01, 0.1, 0.5, 1.0]                         & Shrinks the contribution of each classifier. Smaller values require more stages to achieve the same training error. \\ \hline
\multirow{3}{*}{LogisticRegression}         & C                       & [0.001, 0.01, 0.1, 1, 10, 100]                & Inverse of regularization strength. Smaller values specify stronger regularization. \\ \cline{2-4} 
                                 & penalty                 & ['l1', 'l2', 'elasticnet']                    & Norm used in the penalization. Helps in controlling overfitting by penalizing large coefficients. \\ \cline{2-4} 
                                 & solver                  & ['liblinear', 'saga']                         & Algorithm to use in the optimization problem. Different solvers have different strengths and computational efficiencies. \\ \hline
\multirow{6}{*}{RandomForest}     & n\_estimators            & [100, 200, 500, 1000]                         & Number of trees in the forest. More trees can improve performance but also increase computation time. \\ \cline{2-4} 
                                 & max\_features            & ['auto', 'sqrt', 'log2']                      & Number of features to consider when looking for the best split. Helps in reducing overfitting by ensuring diverse trees. \\ \cline{2-4} 
                                 & max\_depth               & [10, 20, 30, None]                            & Maximum depth of the tree. Limits the number of nodes in the tree to prevent overfitting. \\ \cline{2-4} 
                                 & min\_samples\_split      & [2, 5, 10]                                    & Minimum number of samples required to split an internal node. Ensures splits are meaningful by requiring a sufficient number of samples. \\ \cline{2-4} 
                                 & min\_samples\_leaf       & [1, 2, 4]                                     & Minimum number of samples required to be at a leaf node. Ensures leaves have enough samples to be statistically meaningful. \\ \cline{2-4} 
                                 & bootstrap               & [True, False]                                 & Whether bootstrap samples are used when building trees. Affects the variance and bias of the model. \\ \hline
\multirow{5}{*}{DecisionTree}     & max\_features            & ['auto', 'sqrt', 'log2']                      & Number of features to consider when looking for the best split. Helps in reducing overfitting. \\ \cline{2-4} 
                                 & max\_depth               & [10, 20, 30, None]                            & Maximum depth of the tree. Limits the number of nodes in the tree to prevent overfitting. \\ \cline{2-4} 
                                 & min\_samples\_split      & [2, 5, 10]                                    & Minimum number of samples required to split an internal node. Ensures splits are meaningful by requiring a sufficient number of samples. \\ \cline{2-4} 
                                 & min\_samples\_leaf       & [1, 2, 4]                                     & Minimum number of samples required to be at a leaf node. Ensures leaves have enough samples to be statistically meaningful. \\ \cline{2-4} 
                                 & criterion               & ['gini', 'entropy']                           & Function to measure the quality of a split. Different criteria can lead to different splits and tree structures. \\ \bottomrule
\end{tabular}
\caption{Hyperparameter grids for RandomizedSearchCV used for the classifier comparison on the NS dataset. Explanations are from the \href{https://scikit-learn.org/stable/modules/generated/sklearn.model_selection.RandomizedSearchCV.html}{Scikit Learn Website}.}
\label{tab:params}
\end{table}\newpage

\section{Confusion Matrices}
\subsection{US Air data} \label{sec:cm_us}
\begin{figure}[H]
    \centering
    \begin{subfigure}[t]{0.19\linewidth}
        \centering
        \includegraphics[width=\linewidth]{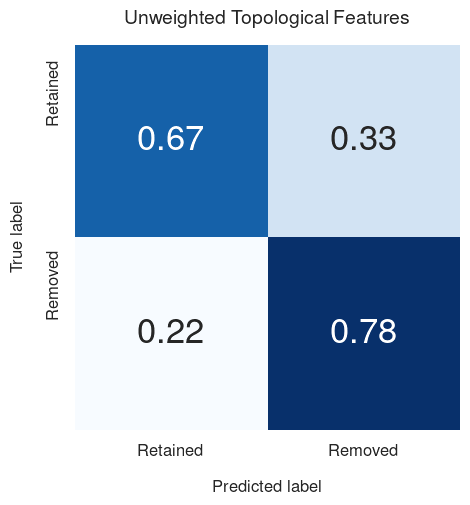}
    \end{subfigure}
    \hfill
    \begin{subfigure}[t]{0.19\linewidth}
        \centering
        \includegraphics[width=\linewidth]{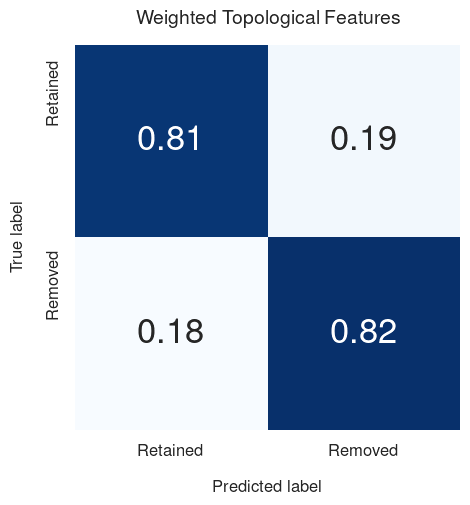}
    \end{subfigure}
    \hfill
    \begin{subfigure}[t]{0.19\linewidth}
        \centering
        \includegraphics[width=\linewidth]{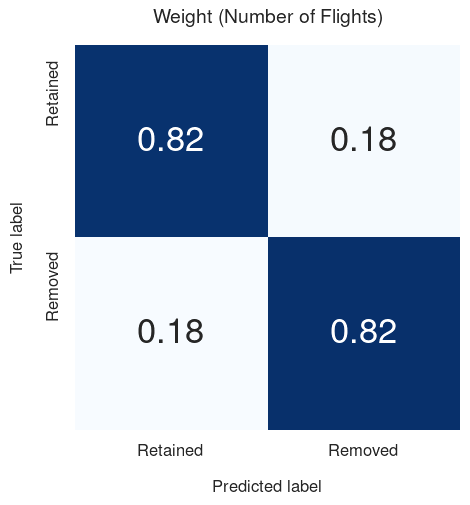}
    \end{subfigure}
    \hfill
    \begin{subfigure}[t]{0.19\linewidth}
        \centering
        \includegraphics[width=\linewidth]{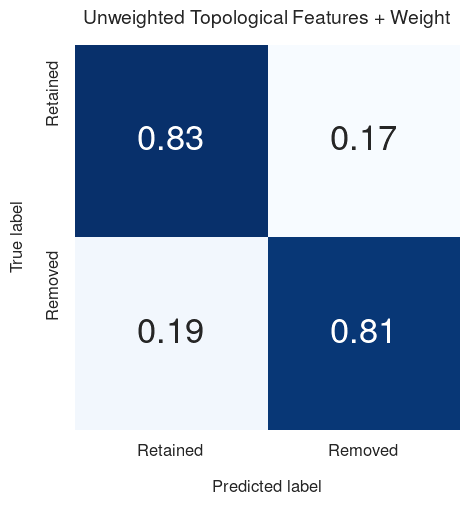}
    \end{subfigure}
    \hfill
    \begin{subfigure}[t]{0.19\linewidth}
        \centering
        \includegraphics[width=\linewidth]{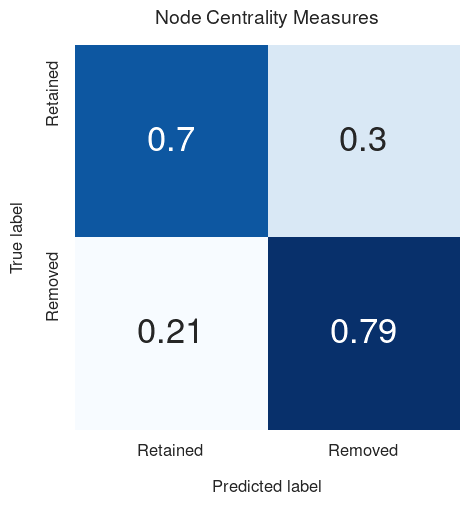}
    \end{subfigure}
    \hfill
    \caption{Confusion matrices for simultaneous testing on the US air data with different feature sets.}
    \label{fig:us_cm_sim}
\end{figure}

\begin{figure}[H]
    \centering
    \begin{subfigure}[t]{0.19\linewidth}
        \centering
        \includegraphics[width=\linewidth]{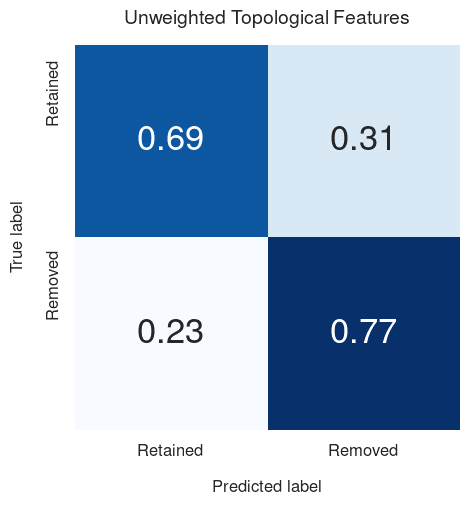}
    \end{subfigure}
    \hfill
    \begin{subfigure}[t]{0.19\linewidth}
        \centering
        \includegraphics[width=\linewidth]{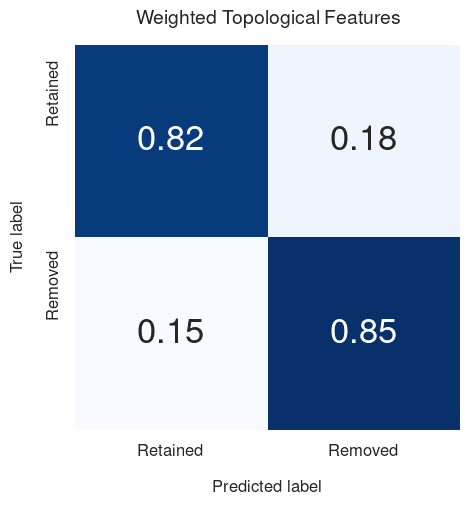}
    \end{subfigure}
    \hfill
    \begin{subfigure}[t]{0.19\linewidth}
        \centering
        \includegraphics[width=\linewidth]{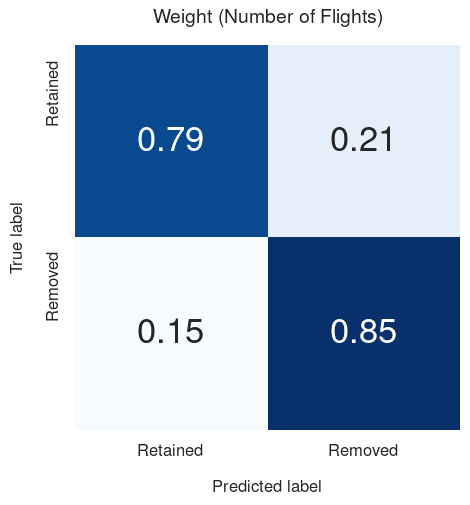}
    \end{subfigure}
    \hfill
    \begin{subfigure}[t]{0.19\linewidth}
        \centering
        \includegraphics[width=\linewidth]{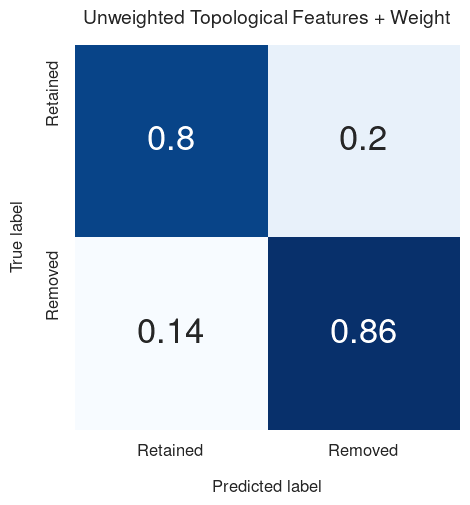}
    \end{subfigure}
    \hfill
    \begin{subfigure}[t]{0.19\linewidth}
        \centering
        \includegraphics[width=\linewidth]{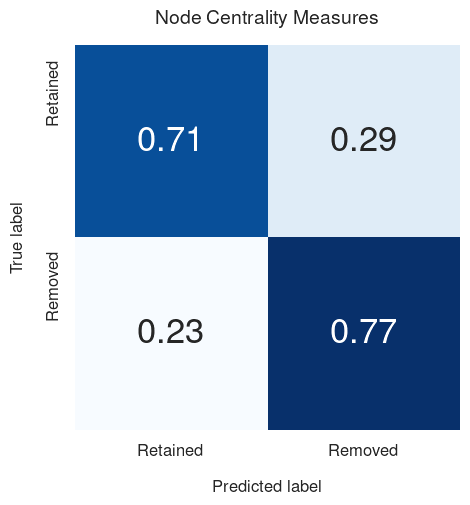}
    \end{subfigure}
    \hfill
    \caption{Confusion matrices for non-simultaneous testing on the US air data with different feature sets.}
    \label{fig:us_cm_nonsim}
\end{figure} \newpage

\subsection{NS data} \label{sec:ns_cm}
\begin{figure}[H]
    \centering
    \begin{subfigure}[t]{0.32\linewidth}
        \centering
        \includegraphics[width=\linewidth]{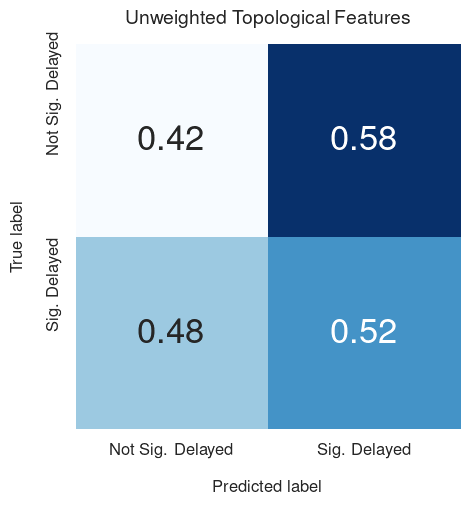}
    \end{subfigure}
    \hfill
    \begin{subfigure}[t]{0.32\linewidth}
        \centering
        \includegraphics[width=\linewidth]{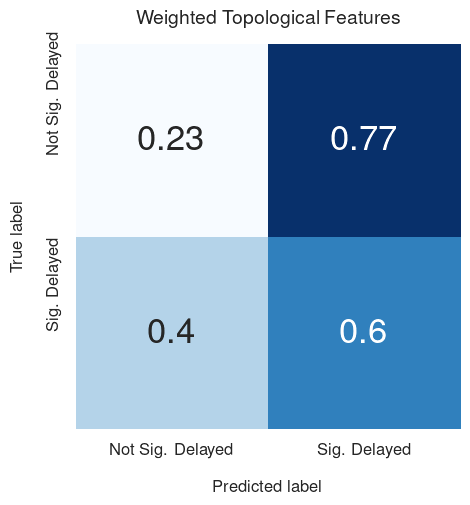}
    \end{subfigure}
    \hfill
    \begin{subfigure}[t]{0.32\linewidth}
        \centering
        \includegraphics[width=\linewidth]{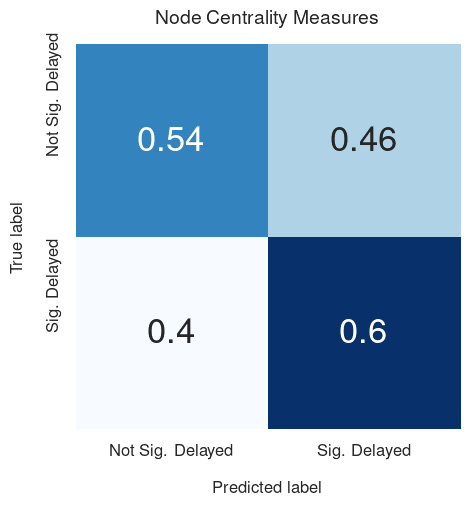}
    \end{subfigure}
    \hfill
    \caption{Confusion matrices for simultaneous testing on the NS data with different feature sets.}
    \label{fig:nl_cm_sim}
\end{figure}

\begin{figure}[H]
    \centering
    \begin{subfigure}[t]{0.32\linewidth}
        \centering
        \includegraphics[width=\linewidth]{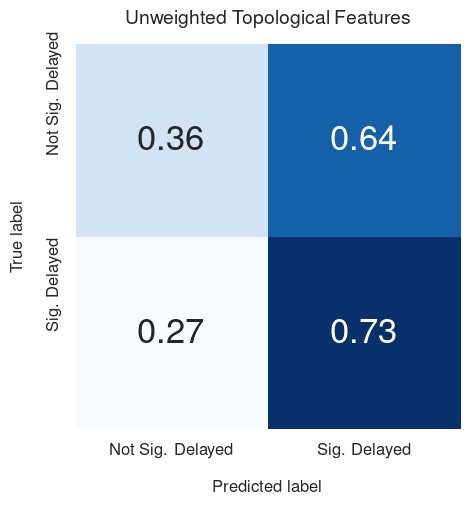}
    \end{subfigure}
    \hfill
    \begin{subfigure}[t]{0.32\linewidth}
        \centering
        \includegraphics[width=\linewidth]{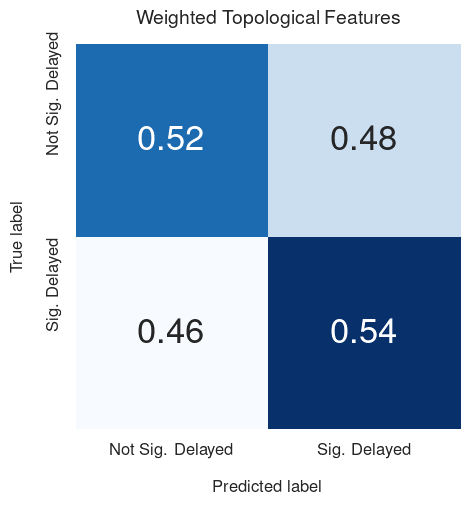}
    \end{subfigure}
    \hfill
    \begin{subfigure}[t]{0.32\linewidth}
        \centering
        \includegraphics[width=\linewidth]{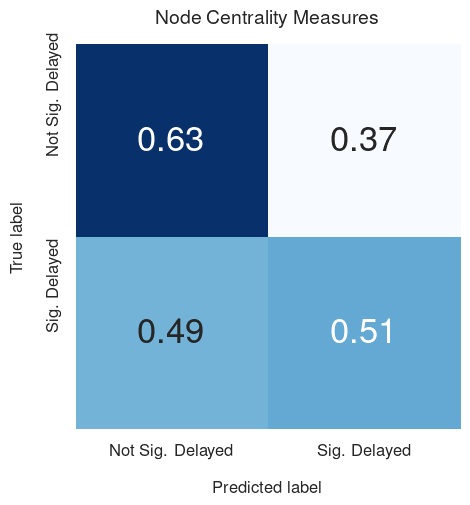}
    \end{subfigure}
    \hfill
    \caption{Confusion matrices for non-simultaneous testing on the NS data with different feature sets.}
    \label{fig:nl_cm_nonsim}
\end{figure}\newpage

\section{SHAP Values Features for NS Data} \label{sec:ns_shap}
\begin{figure}[H]
    \centering
    \begin{subfigure}[t]{0.49\linewidth}
        \centering
        \includegraphics[width=\linewidth]{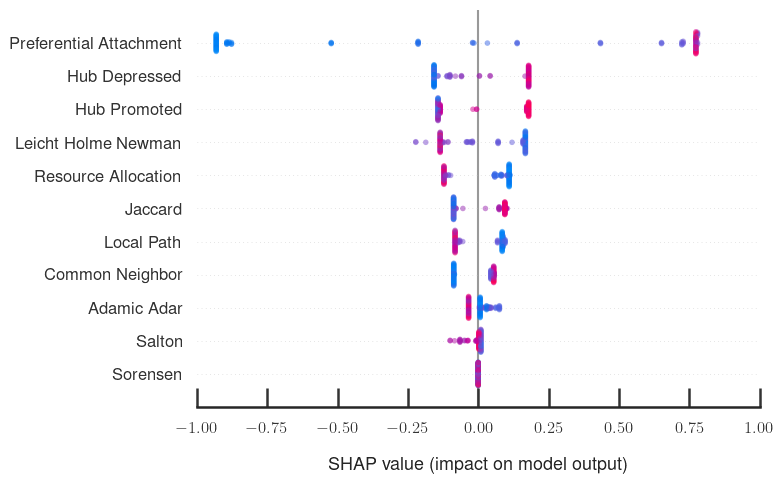}
    \end{subfigure}
    \hfill
    \begin{subfigure}[t]{0.49\linewidth}
        \centering
        \includegraphics[width=\linewidth]{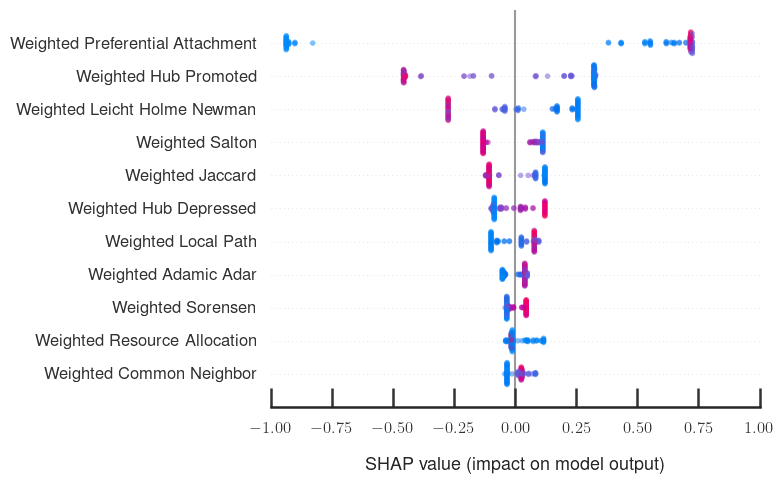}
    \end{subfigure}
    \begin{subfigure}[t]{0.49\linewidth}
        \centering
        \includegraphics[width=\linewidth]{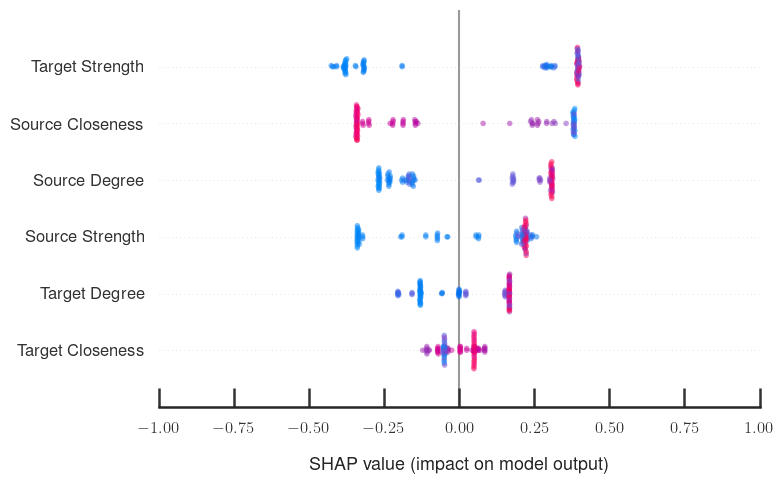}
    \end{subfigure}
    \hfill
    \caption{SHAP figures for the three different feature sets used as predictors in the NS data.}
    \label{fig:nl_shap}
\end{figure}\newpage

\section{Metrics} \label{sec:metrics}
\subsection{Simultaneous Testing}
\begin{table}[H]
\centering
\begin{minipage}{0.48\textwidth}
\centering
\begin{tabular}{p{3.7cm}p{1cm}p{1cm}p{1cm}}
\toprule
\textbf{Classifier} & \textbf{TF} & \textbf{WTF} & \textbf{NCM} \\
\midrule
AdaBoost & 0.627 & 0.627 & 0.639 \\ 
DecisionTree & 0.597 & 0.601 & 0.597 \\ 
GradientBoosting & 0.625 & 0.623 & 0.629 \\ 
LogisticRegression & 0.637 & \textbf{0.640} & 0.637 \\ 
RandomForest & 0.625 & 0.629 & \textbf{0.648} \\ 
XGBoost & \textbf{0.638} & 0.632 & 0.647 \\ 
\bottomrule
\end{tabular}
\caption{F1 scores of classifiers for every feature set when simultaneous testing.}
\label{tab:classifier_f1_scores}
\end{minipage}
\hfill
\begin{minipage}{0.48\textwidth}
\centering
\begin{tabular}{p{3.7cm}p{1cm}p{1cm}p{1cm}}
\toprule
\textbf{Classifier} & \textbf{TF} & \textbf{WTF} & \textbf{NCM} \\
\midrule
AdaBoost & 0.646 & 0.653 & 0.661 \\ 
DecisionTree & 0.615 & 0.614 & 0.595 \\ 
GradientBoosting & 0.645 & 0.649 & 0.665 \\ 
LogisticRegression & \textbf{0.688} & 0.685 & 0.691 \\ 
RandomForest & 0.666 & 0.671 & 0.695 \\ 
XGBoost & 0.684 & \textbf{0.686} & \textbf{0.701} \\ 
\bottomrule
\end{tabular}
\caption{ROC AUC scores of classifiers for every feature set when simultaneous testing.}
\label{tab:classifier_roc_auc_scores}
\end{minipage}
\end{table}

\subsection{Non-Simultaneous Testing}
\begin{table}[H]
\centering
\begin{minipage}{0.48\textwidth}
\centering
\begin{tabular}{p{3.7cm}p{1cm}p{1cm}p{1cm}}
\toprule
\textbf{Classifier} & \textbf{TF} & \textbf{WTF} & \textbf{NCM} \\
\midrule
AdaBoost & \textbf{0.501} & 0.510 & 0.513 \\ 
DecisionTree & 0.481 & 0.485 & 0.522 \\ 
GradientBoosting & 0.506 & 0.513 & \textbf{0.525} \\ 
LogisticRegression & 0.478 & \textbf{0.520} & 0.502 \\ 
RandomForest & 0.497 & 0.510 & 0.519 \\ 
XGBoost & 0.484 & 0.500 & 0.494 \\ 
\bottomrule
\end{tabular}
\caption{F1 scores of classifiers for every feature set when non-simultaneous testing.}
\label{tab:non_simultaneous_f1_scores}
\end{minipage}
\hfill
\begin{minipage}{0.48\textwidth}
\centering
\begin{tabular}{p{3.7cm}p{1cm}p{1cm}p{1cm}}
\toprule
\textbf{Classifier} & \textbf{TF} & \textbf{WTF} & \textbf{NCM} \\
\midrule
AdaBoost & 0.533 & 0.537 & 0.548 \\ 
DecisionTree & 0.522 & 0.524 & 0.537 \\ 
GradientBoosting & 0.529 & 0.537 & 0.559 \\ 
LogisticRegression & 0.535 & 0.552 & 0.552 \\ 
RandomForest & \textbf{0.538} & \textbf{0.548} & \textbf{0.569} \\ 
XGBoost & \textbf{0.538} & 0.543 & 0.551 \\ 
\bottomrule
\end{tabular}
\caption{ROC AUC scores of classifiers for every feature set when non-simultaneous testing.}
\label{tab:non_simultaneous_roc_auc_scores}
\end{minipage}
\end{table}

\end{document}